\documentclass[sigconf]{aamas} 
\usepackage[utf8]{inputenc}
\usepackage{amsfonts}
\usepackage{amsmath}
\usepackage{dirtytalk}
\usepackage{subcaption}
\usepackage{graphicx}
\usepackage{balance}
\usepackage{scalerel}
\usepackage{relsize}
\usepackage{stfloats}
\usepackage{subcaption}

\setcopyright{ifaamas}
\acmConference[AAMAS '23]{Proc.\@ of the 22nd International Conference
on Autonomous Agents and Multiagent Systems (AAMAS 2023)}{May 29 -- June 2, 2023}
{London, United Kingdom}{A.~Ricci, W.~Yeoh, N.~Agmon, B.~An (eds.)}
\copyrightyear{2023}
\acmYear{2023}
\acmDOI{}
\acmPrice{}
\acmISBN{}

\acmSubmissionID{569}

\title[AAMAS-2023 Formatting Instructions]{Permutation-Invariant Set Autoencoders with Fixed-Size Embeddings for Multi-Agent Learning}

\author{Ryan Kortvelesy}
\affiliation{
  \institution{University of Cambridge}
  \city{Cambridge}
  \country{United Kingdom}}
\email{rk627@cam.ac.uk}

\author{Steven Morad}
\affiliation{
  \institution{University of Cambridge}
  \city{Cambridge}
  \country{United Kingdom}}
\email{sm2558@cam.ac.uk}

\author{Amanda Prorok}
\affiliation{
  \institution{University of Cambridge}
  \city{Cambridge}
  \country{United Kingdom}}
\email{asp45@cam.ac.uk}

\begin{abstract}

The problem of permutation-invariant learning over set representations is particularly relevant in the field of multi-agent systems---a few potential applications include unsupervised training of aggregation functions in graph neural networks (GNNs), neural cellular automata on graphs, and prediction of scenes with multiple objects. Yet existing approaches to set encoding and decoding tasks present a host of issues, including non-permutation-invariance, fixed-length outputs, reliance on iterative methods, non-deterministic outputs, computationally expensive loss functions, and poor reconstruction accuracy. In this paper we introduce a Permutation-Invariant Set Autoencoder (PISA), which tackles these problems and produces encodings with significantly lower reconstruction error than existing baselines. PISA also provides other desirable properties, including a similarity-preserving latent space, and the ability to insert or remove elements from the encoding. After evaluating PISA against baseline methods, we demonstrate its usefulness in a multi-agent application. Using PISA as a subcomponent, we introduce a novel GNN architecture which serves as a generalised communication scheme, allowing agents to use communication to gain full observability of a system.

\end{abstract}


\begin{document}
\pagestyle{fancy}
\fancyhead{}
\maketitle

\begin{figure*}[t]
    \begin{subfigure}[b]{.49\textwidth}
    	\centering
            \includegraphics[width=0.98\textwidth]{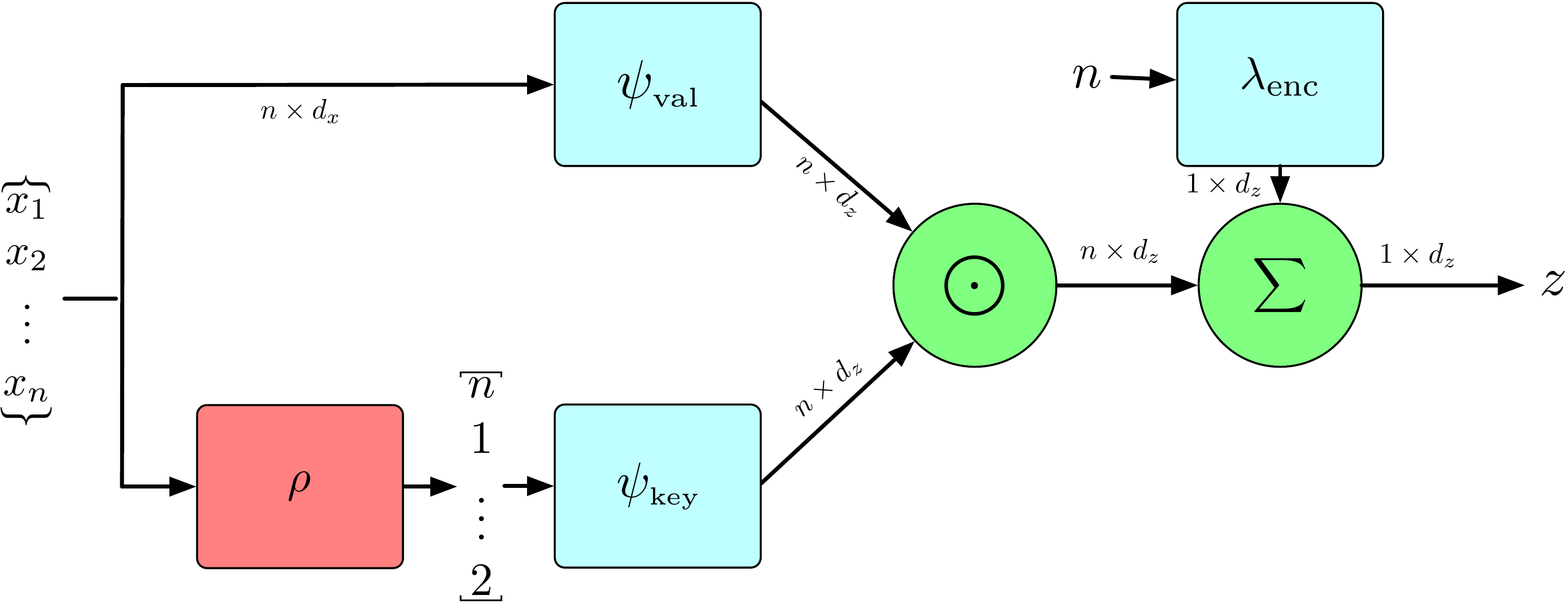}
            \vspace{-0.7mm}
    	\caption{Set Encoder}
    	\label{1a}
    \end{subfigure}
    \begin{subfigure}[b]{.49\textwidth}
            \centering
            \includegraphics[width=0.98\textwidth]{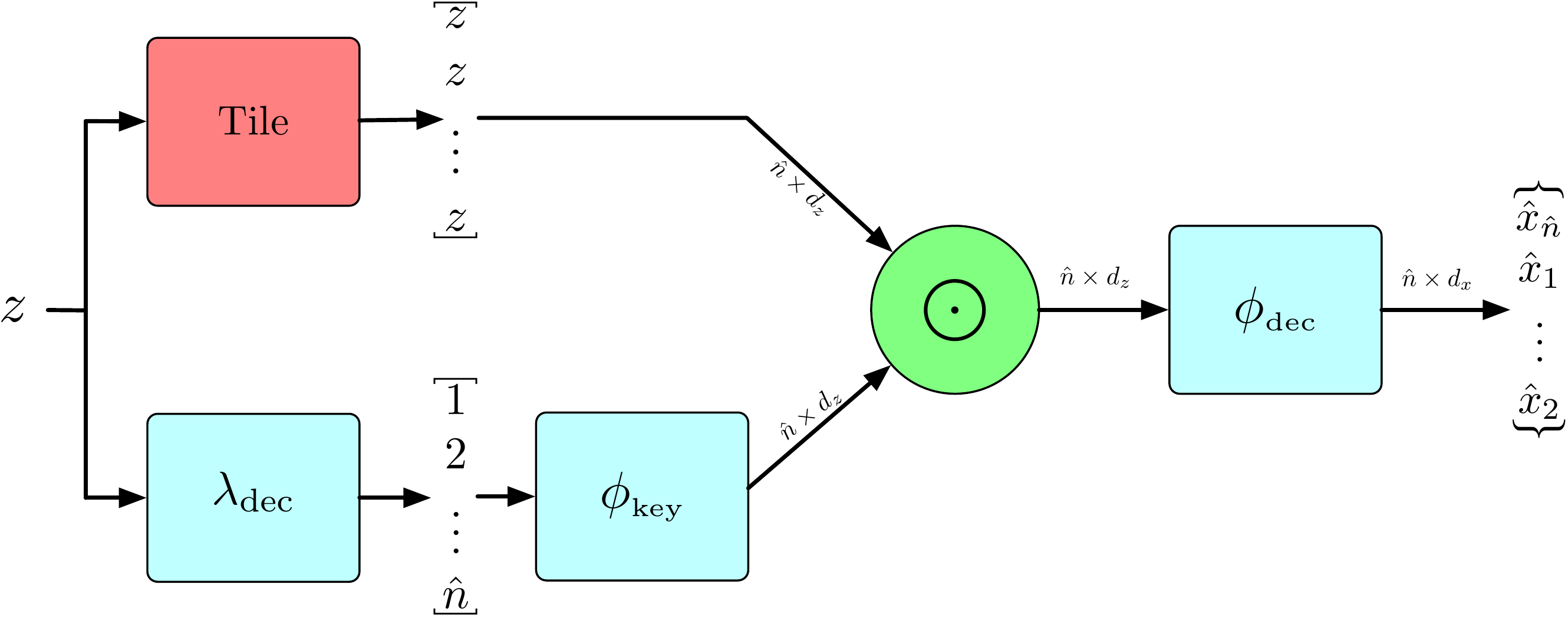}
    	\caption{Set Decoder}
    	\label{1b}
    \end{subfigure}
    \caption{The full set autoencoder architecture. In this schema, the blue modules are learned networks, the green modules are mathematical operations (sum and element-wise product), and the red modules are other non-learned operations. (\subref{1a}) The encoder first assigns keys to values according to a deterministic criterion $\rho$. Then, it encodes the values with $\psi_\mathrm{val}$ and the keys with $\psi_\mathrm{key}$, mapping both to $\mathbb{R}^{d_z}$. Each element is inserted by adding the element-wise product of the encoded key and value to the latent state. The cardinality of the set is also encoded by adding $\lambda_\mathrm{enc}(n)$ to the embedding. (\subref{1b}) The decoder first predicts the cardinality of the set with $\lambda_\mathrm{dec}$. Next, it produces $\hat{n}$ keys in the same manner as the encoder, and transforms them into queries in $\mathbb{R}^{d_z}$ with $\phi_\mathrm{key}$. Each of these queries is element-wise multiplied by the hidden state to produce an element-specific encoding. Finally, $\phi_\mathrm{dec}$ transforms these encodings into the reconstructed set.}
    \label{fig:sae architecture}
\end{figure*}

\section{Introduction}

Over time, machine learning research has placed an increasing emphasis on utilising relational inductive biases \cite{relational}. By focusing on the underlying relationships in graph structured data, it has become possible to create models with superior performance and generalisation. These models are particularly useful in multi-agent learning, where most data is structured as a graph. For example, graph neural networks (GNNs) are used to learn communication amongst a variable number of neighbours \cite{qingbiaognn, modgnn}, graph pooling is used to facilitate value factorisation in a variable-sized system of agents 
\cite{qmix, qgnn}, and set encoders are used to classify a variable number of observations (such as pointclouds) \cite{deepsets, pointnet}.

While there are many architectures suited for encoding variable-sized sets of elements, there is a lack of methods that can be used to \textit{decode} sets. Consequently, tasks that require models that can encode \textit{and} decode sets represent an underexplored field within multi-agent learning. One potential application for set autoencoders in this domain is learning encoders and communication schemes in an unsupervised manner. This is particularly useful in multi-agent reinforcement learning (MARL), where the task of learning these functions with just the reward signal represents a significant bottleneck with respect to sample efficiency \cite{learnlatent}.

For a set autoencoder to be useful in a multi-agent context, it must satisfy several properties. First, it must have a low reconstruction error---the primary purpose of the model is to preserve the information being encoded. Second, it must be permutation-invariant. There is no inherent ordering of agents in a system or edges in a graph, so the model should generalise across all possible permutations of inputs. Finally, it should have a similarity-preserving latent space. It is desirable for similar inputs to produce similar embeddings, as it is more robust to noise and it enables generalisation to novel inputs (via interpolation by a continuous function).

Unfortunately, the few existing approaches which tackle the task of set autoencoding exhibit an array of problems, including non-permutation-invariance \cite{rnn, transformers}, fixed-length outputs \cite{dspn}, reliance on iterative methods \cite{dspn}, non-deterministic outputs \cite{tspn}, and computationally expensive loss functions \cite{dspn, tspn}. Even more importantly, these methods exhibit a relatively high reconstruction error in our experiments.

In this paper, we present a novel set autoencoder model which addresses these issues. In contrast to existing methods, our approach operates on the principle of using key-value pairs to encode and decode elements. It also does not require solving an optimisation problem in the loss function which calculates the minimum assignment between input and output elements. We also demonstrate these advantages experimentally, showing that our method can encode and decode sets with a lower reconstruction error than current state-of-the-art methods.

To evaluate our model, we demonstrate its usefulness in two multi-agent learning problems. In our first experiment, we evaluate our set autoencoder against baselines on the task of random set encoding and reconstruction. In the multi-agent domain, a solution to this task constitutes a method for aggregating elements in a manner which minimises the loss of information (which can be used as an aggregation function for neighbours' messages at the agent level, or as a graph pooling layer at the global level). In our second experiment, we tackle the problem of combining the states of agents with partial observability through communication to obtain global observability. Within our solution, we define a novel GNN architecture which leverages our set autoencoder as its internal mechanism.

\textbf{Contributions.}

\begin{itemize}
    \item We present PISA, a novel, permutation-invariant set autoencoder architecture which exhibits significantly lower reconstruction error than existing methods.
    \item Our method preserves similarity in the latent space. We analyse the effect of interpolating in the latent space, showing that our method smoothly transitions between the initial and target sets, while most other methods do not.
    \item Unlike the baselines, our method allows elements to be inserted or removed from the latent state after it has been initially encoded, enabling transforms within the latent space.
    \item We use PISA as a component in a novel GNN which serves as an application-agnostic communication scheme for multi-agent systems. 
\end{itemize}

\section{Related Work}


Permutation-invariant set autoencoders have become popular with the recent development of Deep Sets \cite{deepsets} and to a lesser extent, graph neural networks \cite{gcn} (GNNs). Both are permutation invariant, allowing them to serve as general models for handling variable-sized, unordered data. Methods like Featurewise Sort Pooling (FSPool), Deep Set Prediction Networks (DSPN), and Transformer Set Prediction Networks (TSPN) maintain permutation invariance by using elements from Deep Set theory.


FSPool \cite{zhang_fspool_2020} encodes a set of elements $\mathcal{X} = \{x_1, \dots, x_n\}$ into a single, fixed-size latent representation $z$. It sorts the inputs elements along the feature axis, applies a linear transform, and then sums them together to produce a single embedding $z$ for the set. The authors also propose an FSPool decoder, which maps the embedding $z$ into a prediction of the input set $\hat{\mathcal{X}} = \{\hat{x}_1, \dots, \hat{x}_n\}$. Their decoder unpools $z$ using a linear transformation, and unsorts the unpooled vectors using a differentiable sorting network. FSPool maps $x_i$ to $\hat{x_i}$ via the sort and unsort operations.

The authors of DSPN find that the FSPool decoder is insufficent for solving even simple tasks \cite{dspn}. They use FSPool in the encoder, but replace the decoder with an iterative method. Starting with an initial guess $\hat{\mathcal{X}}^{(0)}$, $\hat{\mathcal{X}}^{(t)}$ is iteratively updated for $T$ iterations, following the gradient of $-\frac{\partial \mathcal{L}(\hat{\mathcal{X}}^{(t)}, z)} {\partial \hat{\mathcal{X}}^{(t)}}$. That is, the output set is updated until its encoding matches a given latent state. Note that because DSPN is an iterative method, its inference speed is slow. Furthermore, it learns a domain-specific initial $\hat{X}^{(0)}$, causing the number of parameters to scale with the size of the largest set. While DSPN can be modified to accommodate variable-size sets by predicting masks, the prediction size is limited by the size of $\hat{X}^{(0)}$. DSPN also requires each $\hat{x}_i$ to be mapped to each $x_i$ in order to compute the training loss. This mapping is computed by the Chamfer algorithm ($O(n^2)$) for large sets, and the Hungarian algorithm ($O(n^3)$) for small sets.


TSPN improves upon DSPN by replacing the iterative decoder with a single pass of a transformer. TSPN samples $\hat{\mathcal{X}}^{(0)}$ from a learned normal distribution, decoupling the number of parameters from the largest set size, but making the predictions non-deterministic. Sampled latent representations $\hat{\mathcal{X}}^{(0)}$ are concatenated along the feature dimension with an embedding $z = \psi(\mathcal{X})$. The resulting set of vectors $\{ \hat{x}^{(0)}_i || \, z \mid i \in [1..n] \}$ are fed to the transformer, which produces the prediction $\hat{\mathcal{X}}$. Similar to DSPN, TSPN uses the Chamfer and Hungarian matching algorithms in its loss function to compute a correspondence between elements in the input set and the predicted set.


Other related work (excluding DSPN, TSPN, and FSPool) tends to break either fixed-size latent representation or permutation invariance constraints. Generalized autoencoders \cite{wang_generalized_2014} project a set of $n$ inputs into a set of $n$ latent states. This precludes their use in many interesting tasks, such as composition with other modules (\textit{e.g.} passing the full set representation through an MLP).




Sequence to sequence models encode ordered sequences of inputs and decode ordered sequences of outputs, and consequently are not permutation invariant. However, prior work \cite{liu20173dcnn} sometimes treats sets as sequences with varying success, utilising models like Gated Recurrent Unit (GRU) \cite{gru}. 

Linear transformers constitute another form of sequence-to-sequence model, which become permutation invariant when the positional encoding is omitted \cite{recurrentfastweight, fastweight}. These approaches map a set to key, value, and query vectors. They produce a fixed-size attention matrix by summing the outer products of each key and value pair. Then, an output is produced by computing the inner product between the attention matrix and query. Unfortunately, this relies on using the keys at decode time, constituting a variable-sized latent state.

All of the methods that we have examined in related work exhibit fundamental problems, including non-permutation-invariance, variable-sized latent states, and reliance on computationally expensive loss functions which introduce instability. Our method addresses all of these issues by tackling the problem of set autoencoding with a significantly different approach.


\section{Method}

\begin{figure*}[t]
    \begin{subfigure}[c]{0.66\textwidth}
        \begin{subfigure}[b]{.49\textwidth}
        	\centering
                \includegraphics[width=0.99\textwidth]{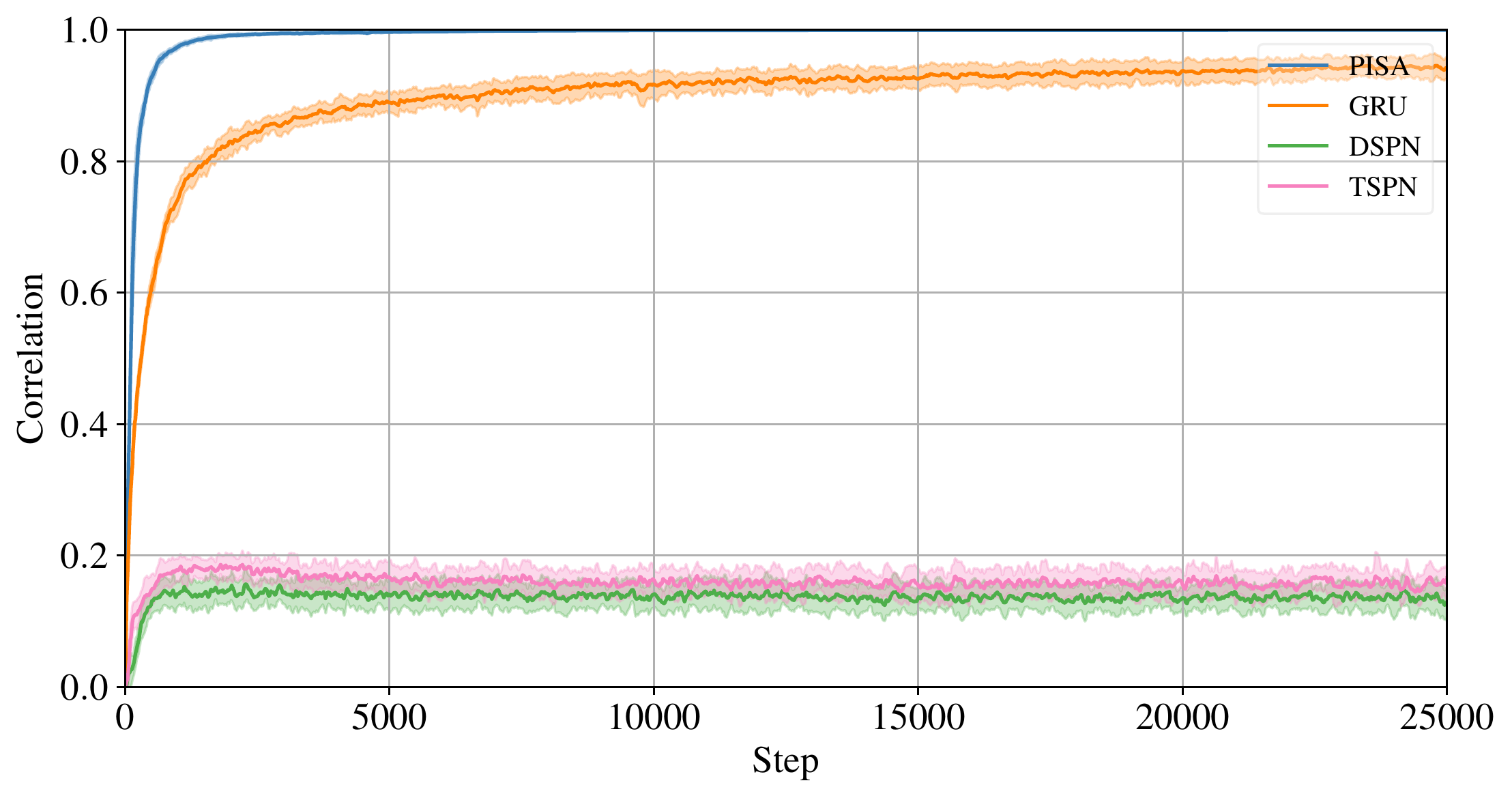}
        	\caption{Correlation ($z \in \mathbb{R}^{96}$)}
                \label{2a}
        \end{subfigure}
        \begin{subfigure}[b]{.49\textwidth}
                \centering
                \includegraphics[width=0.99\textwidth]{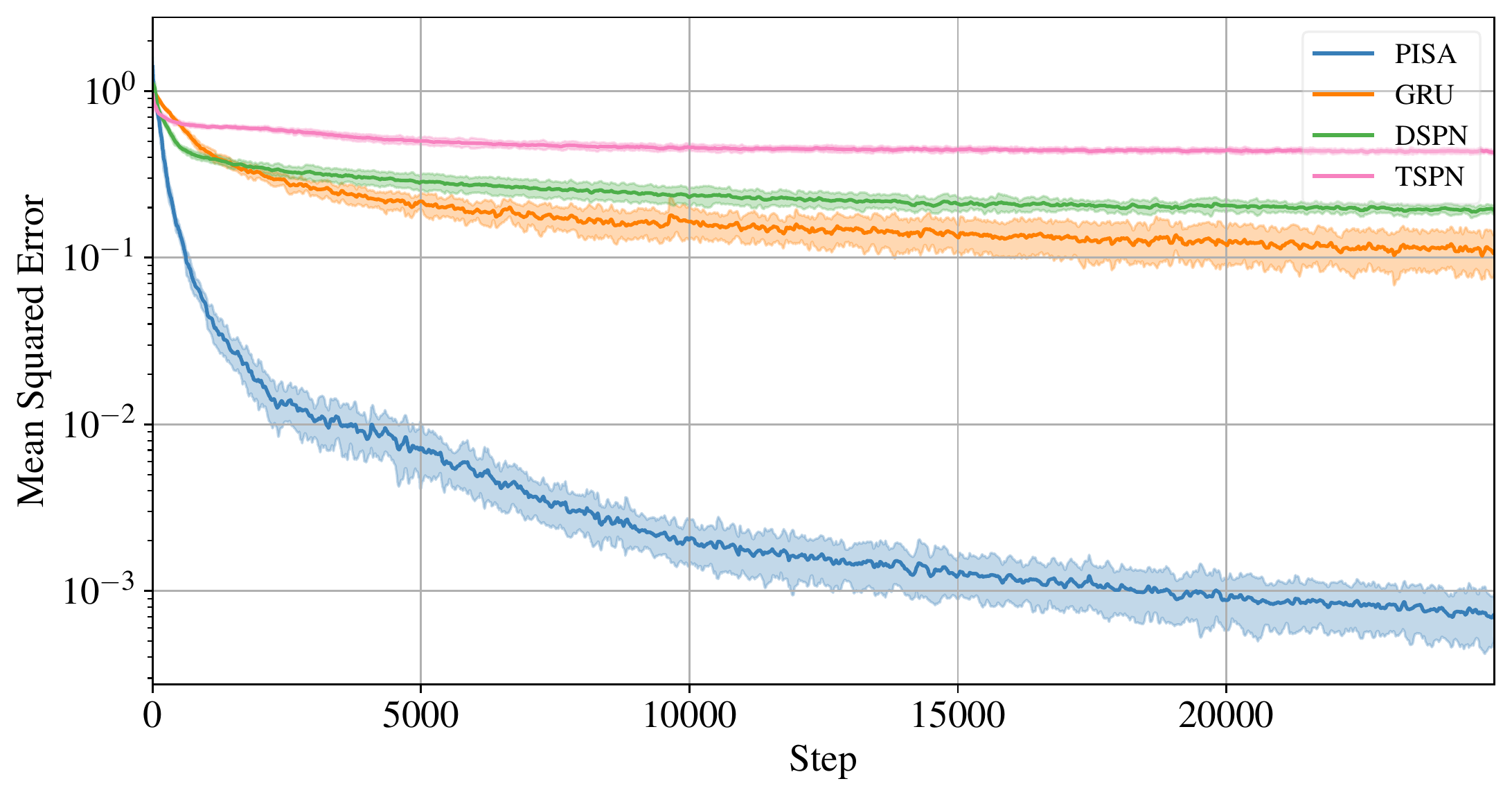}
        	\caption{MSE ($z \in \mathbb{R}^{96}$)}
                \label{2b}
        \end{subfigure}
        \begin{subfigure}[b]{.49\textwidth}
                \centering
                \includegraphics[width=0.99\textwidth]{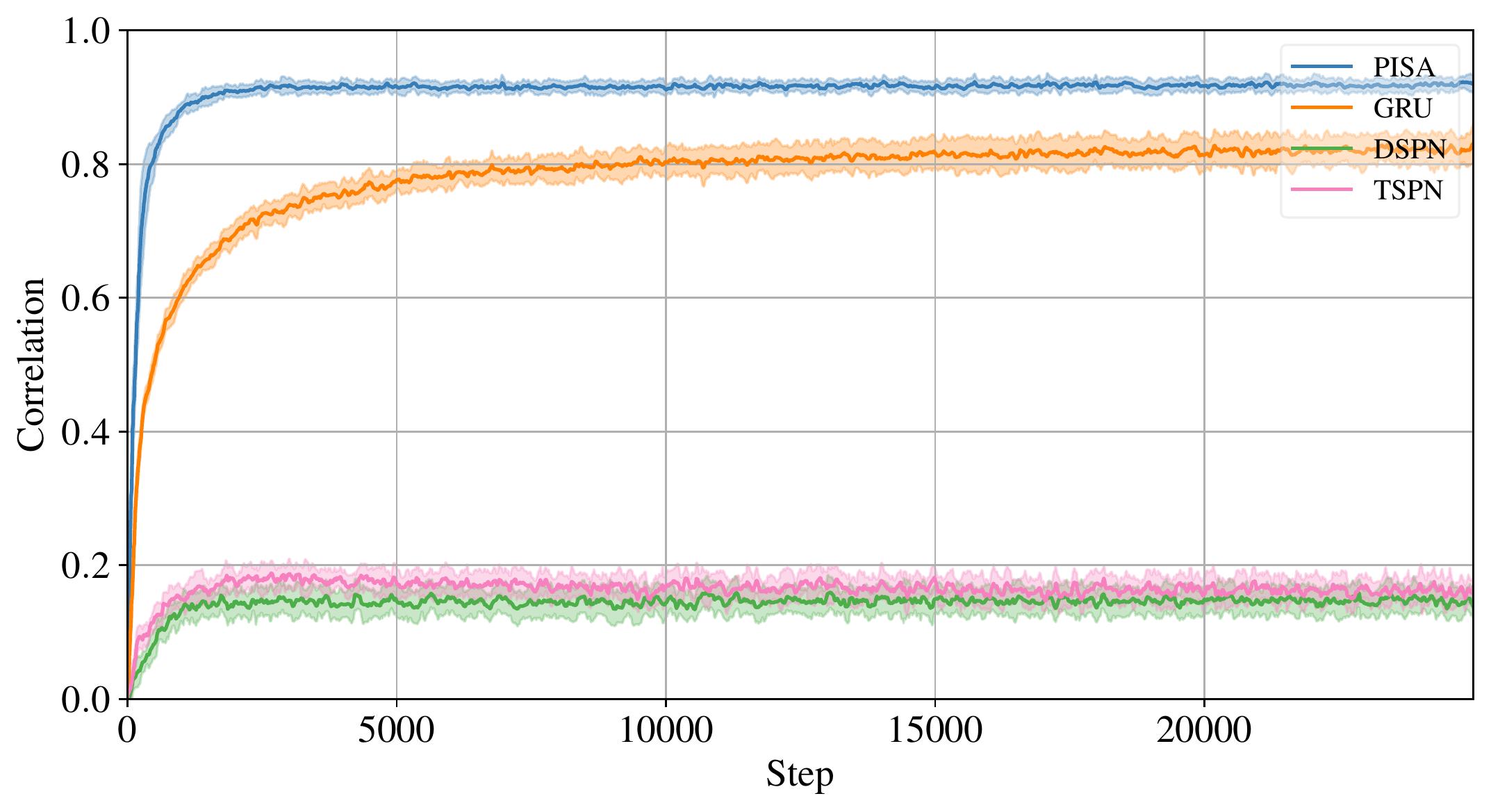}
        	\caption{Correlation ($z \in \mathbb{R}^{48}$)}
                \label{2c}
        \end{subfigure}
        \hfill
        \begin{subfigure}[b]{.49\textwidth}
                \centering
                \includegraphics[width=0.99\textwidth]{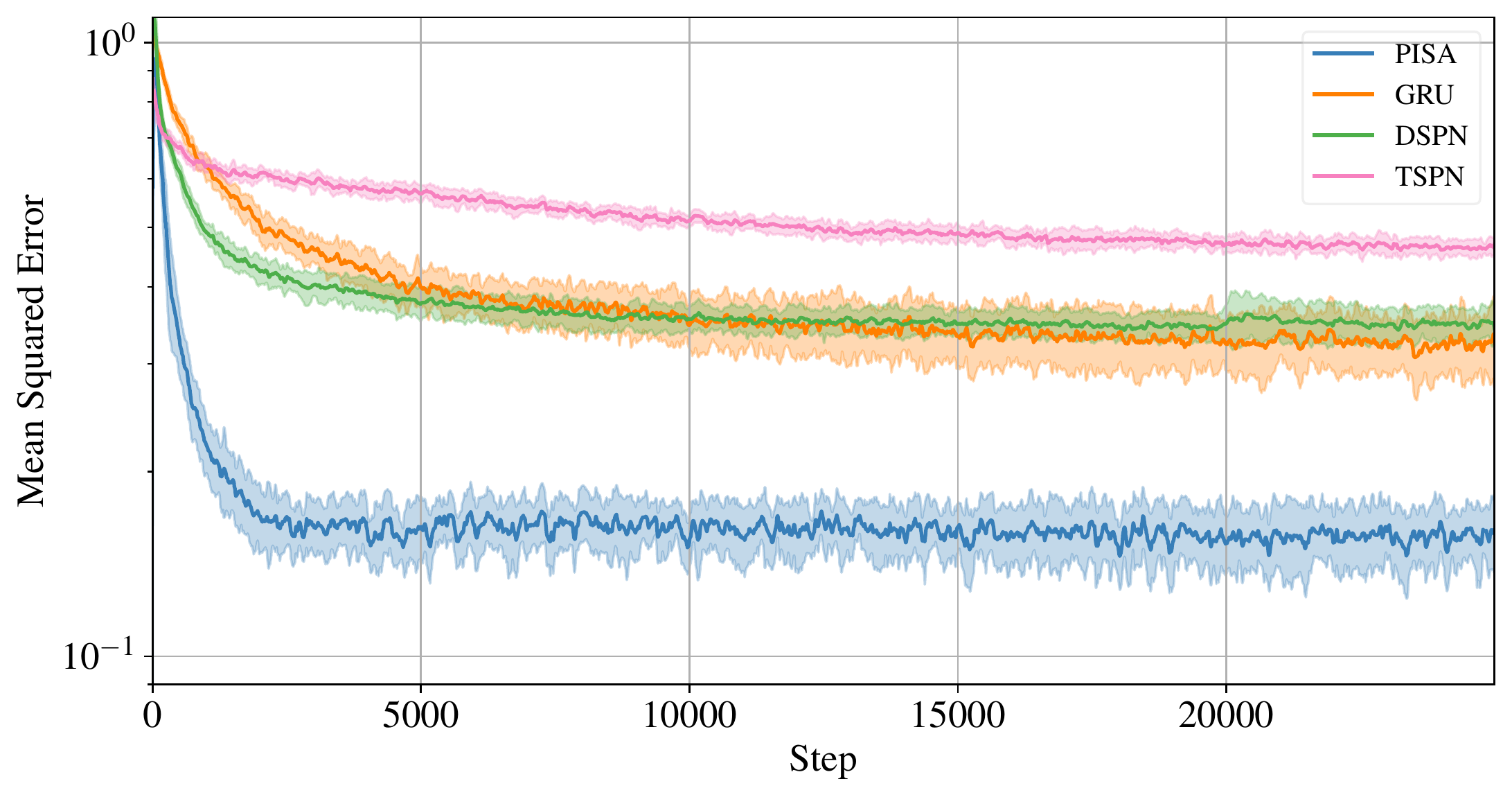}
        	\caption{MSE ($z \in \mathbb{R}^{48}$)}
                \label{2d}
        \end{subfigure}
    \end{subfigure}
    \hfill
    \begin{subfigure}[c]{0.33\textwidth}
        \begin{subfigure}[b]{.49\textwidth}
            \centering
            \includegraphics[width=0.99\textwidth]{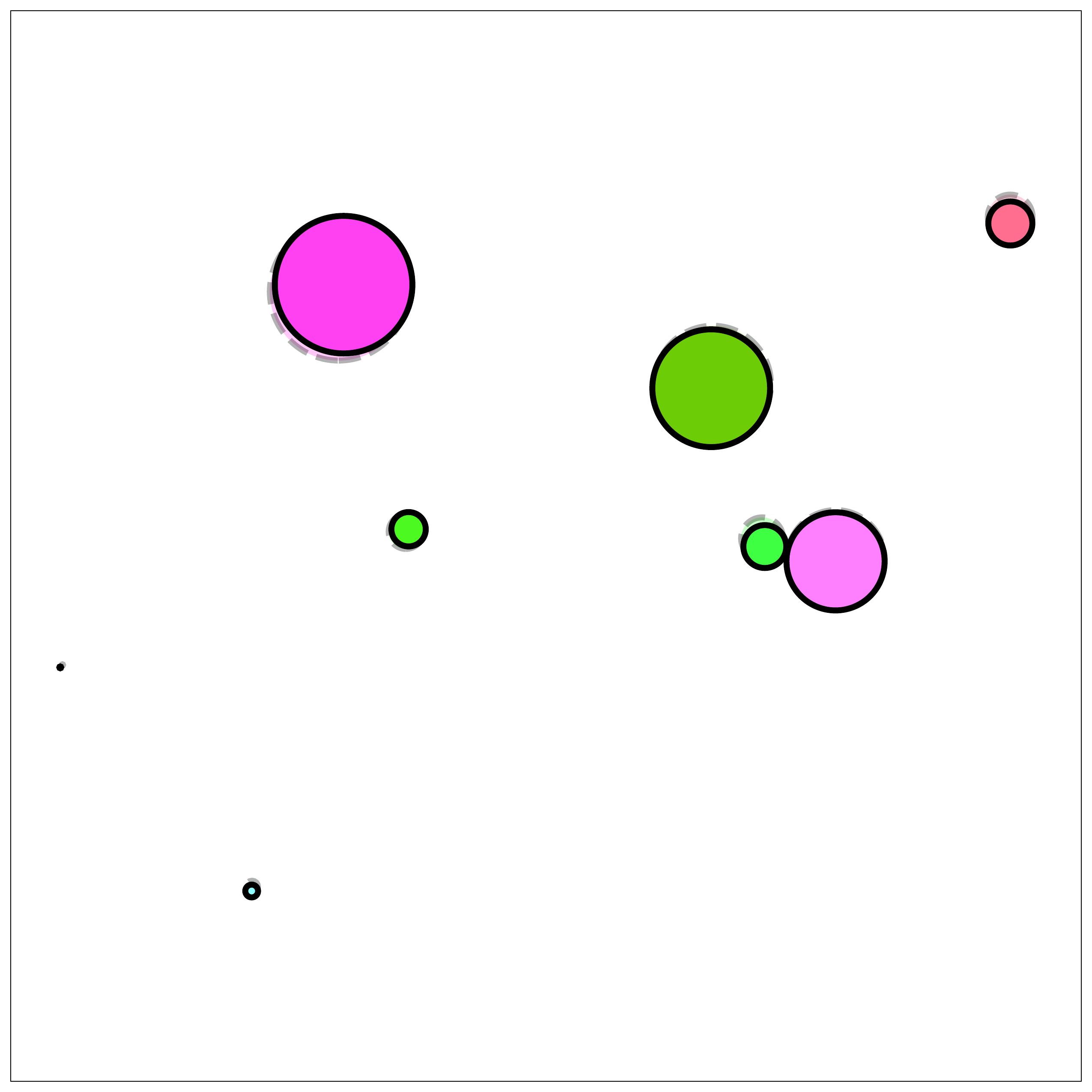}
            \caption{PISA}
            \label{2e}
        \end{subfigure}
        \begin{subfigure}[b]{.49\textwidth}
            \centering
            \includegraphics[width=0.99\textwidth]{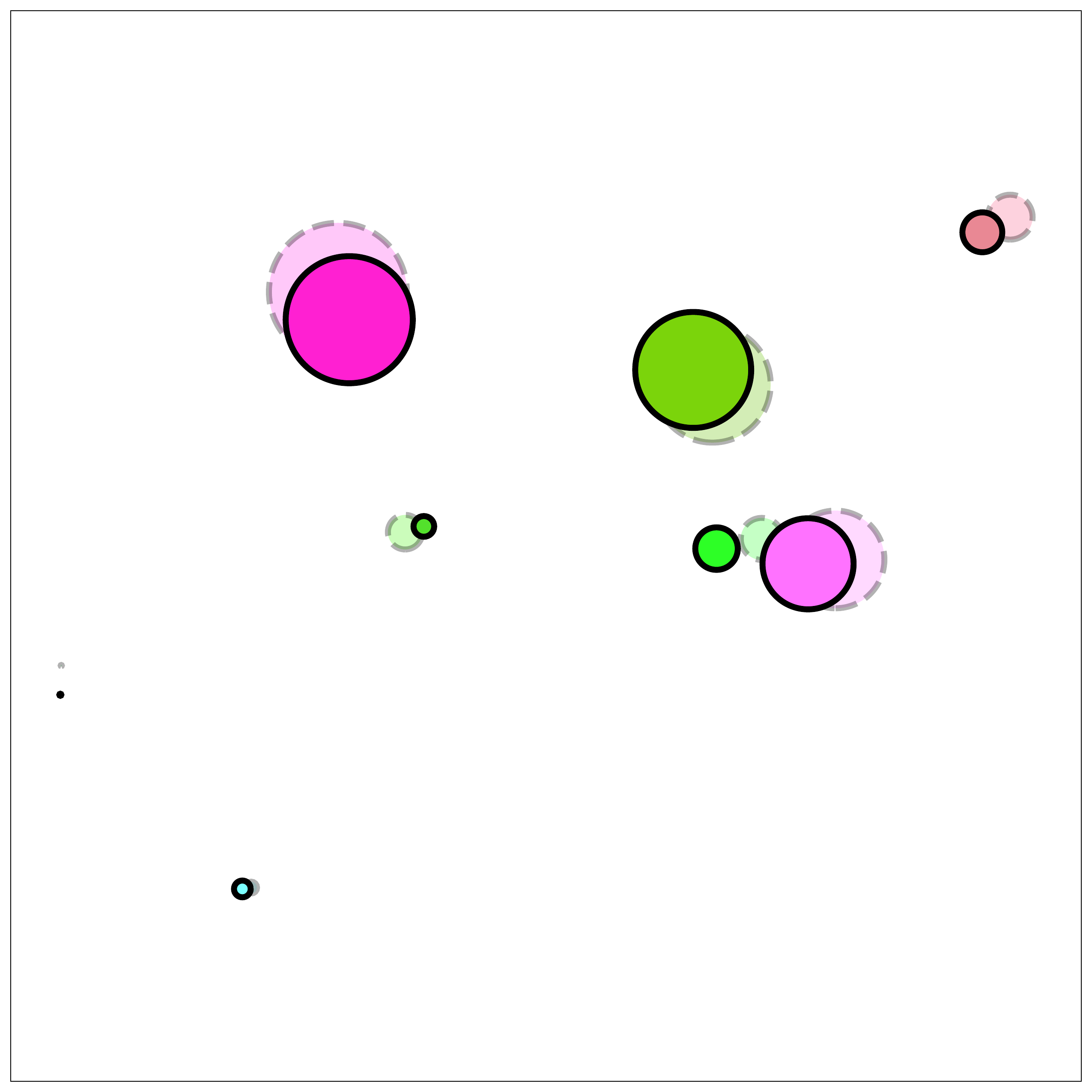}
            \caption{GRU}
            \label{2f}
        \end{subfigure}
        \begin{subfigure}[b]{.49\textwidth}
            \centering
            \includegraphics[width=0.99\textwidth]{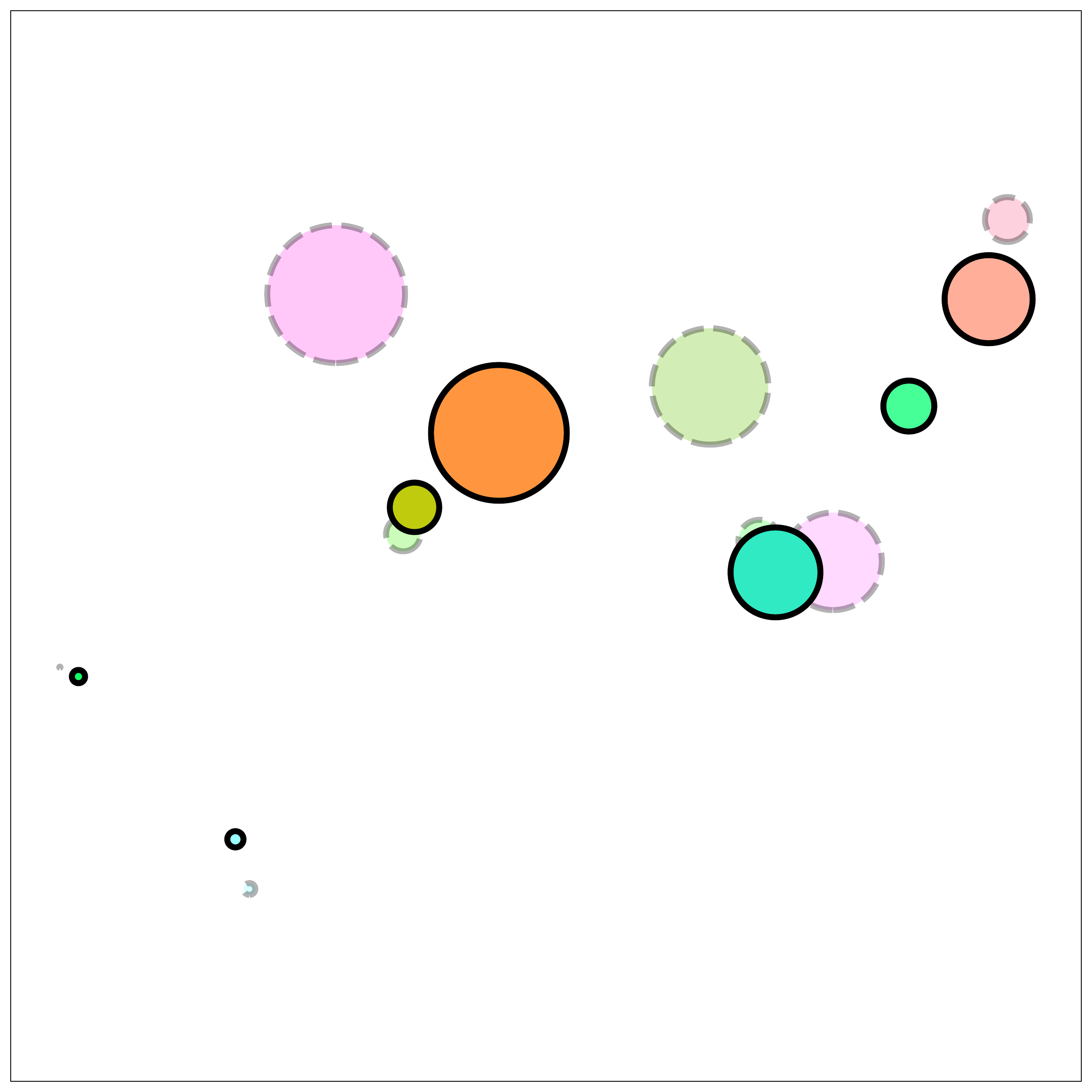}
            \caption{DSPN}
            \label{2g}
        \end{subfigure}
        \begin{subfigure}[b]{.49\textwidth}
            \centering
            \includegraphics[width=0.99\textwidth]{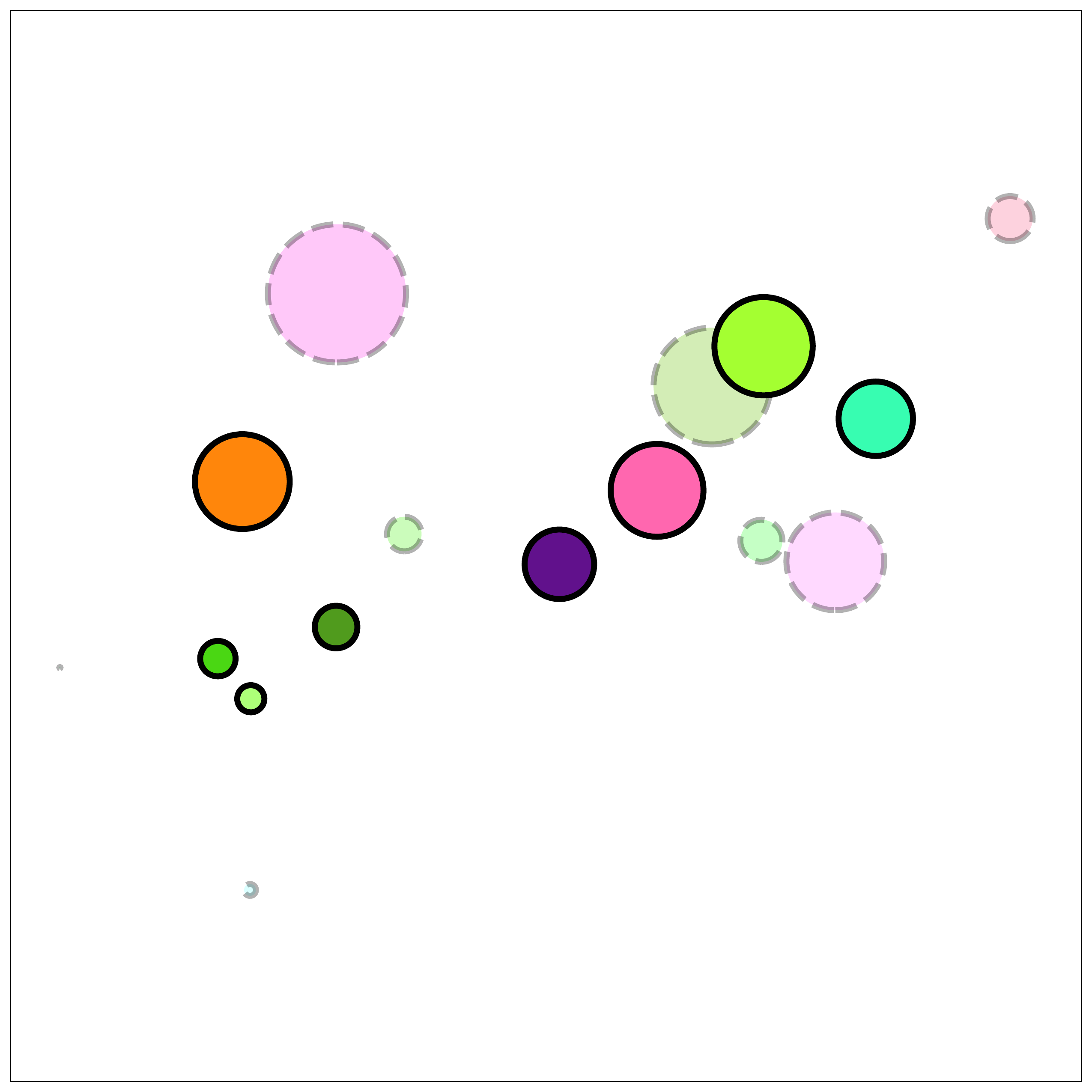}
            \caption{TSPN}
            \label{2h}
        \end{subfigure}
        \vspace{-1mm}
        \label{2b}
    \end{subfigure}
    \caption{Results for the Random Set Reconstruction experiment. (\subref{2a})-(\subref{2d}): Correlation and MSE loss for PISA and baselines with a sufficient latent size ($z \in \mathbb{R}^{96}$) and under compression ($z \in \mathbb{R}^{48}$). (\subref{2e}-\subref{2h}): Reconstruction error for PISA and baselines. The transparent circles represent the ground truth, and the opaque circles represent the prediction.}
    \label{fig:sae results}
\end{figure*}

A set autoencoder architecture consists of an encoder $\psi$ and a decoder $\phi$. The encoder takes a set of elements $\mathcal{X}$ as input
\begin{align}
    \mathcal{X}: \mathbb{R}^{n \times d_x} = \{x_i: \mathbb{R}^{d_x} \mid i \in [1, \dots n]\}
\end{align}
and produces latent state $z$
\begin{align}
    z: \mathbb{R}^{d_z} = \psi(\mathcal{X}),
\end{align}
which is fed to the decoder $\phi$ to predict the original input
\begin{align}
    \hat{\mathcal{X}}: \mathbb{R}^{n \times d_x} = \phi(z).
\end{align}

The goal is to find $\psi$ and $\phi$ such that the \say{set} error between $\mathcal{X}$ and $\hat{\mathcal{X}}$ is minimised (that is, the sizes $|X|$ and $|\hat{X}|$ match, and the MSE error between every corresponding input-output pair of elements is minimised).

Our permutation-invariant set autoencoder (PISA) tackles this problem with a significantly different approach than related works. We introduce a notion of keys and values, popularised by transformers \cite{transformers, fastweight, recurrentfastweight}. Elements are encoded as values, and \say{inserted} into the embedding with keys. By creating key-value pairings, the decoder can extract specific elements with their corresponding queries. This creates a direct correspondence between inputs and outputs, enabling the use of a simple loss function which does not require a matching algorithm to compute the set error.


\begin{figure*}[t]
    \begin{subfigure}[b]{.95\textwidth}
            \centering
            \includegraphics[width=0.135\textwidth]{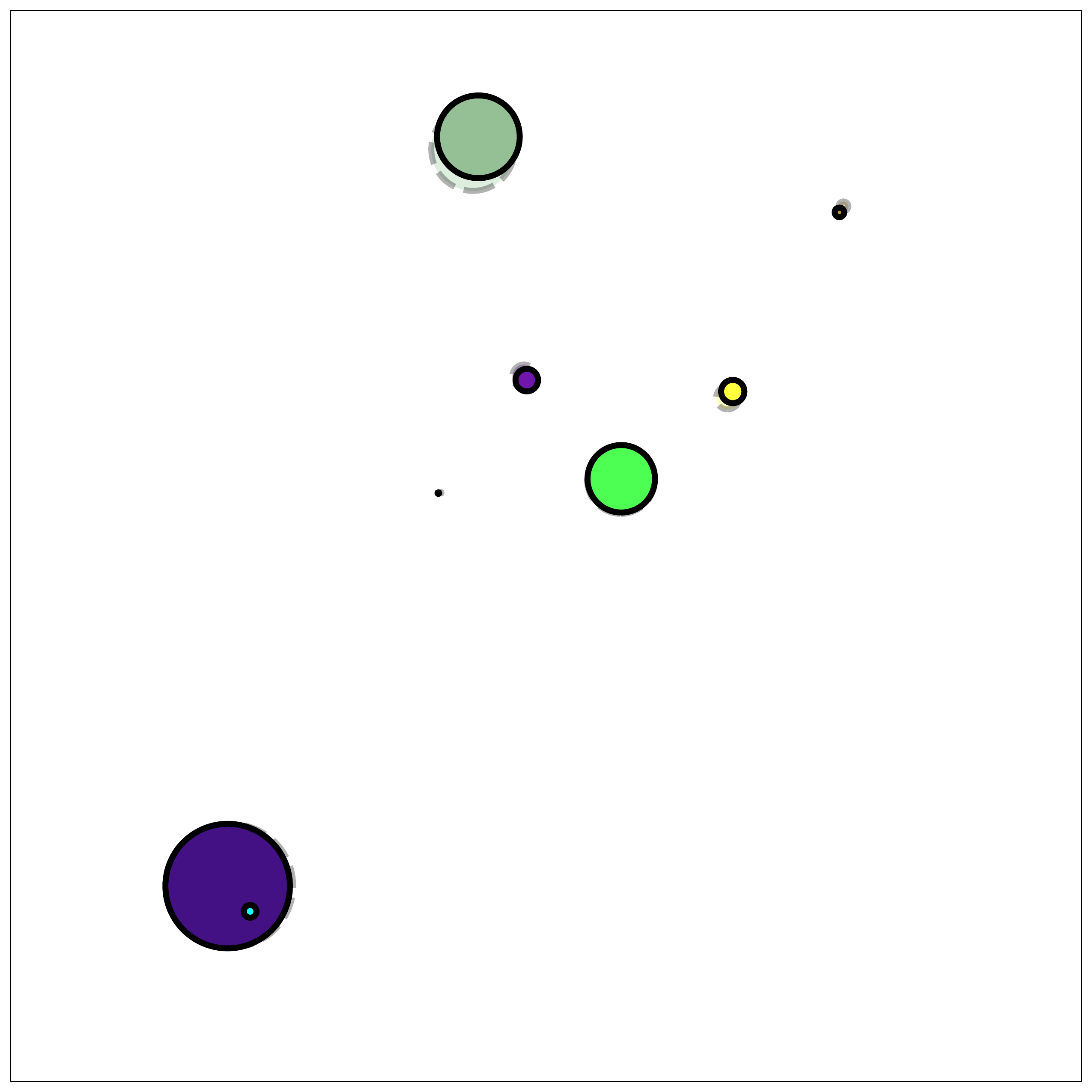}
            \includegraphics[width=0.135\textwidth]{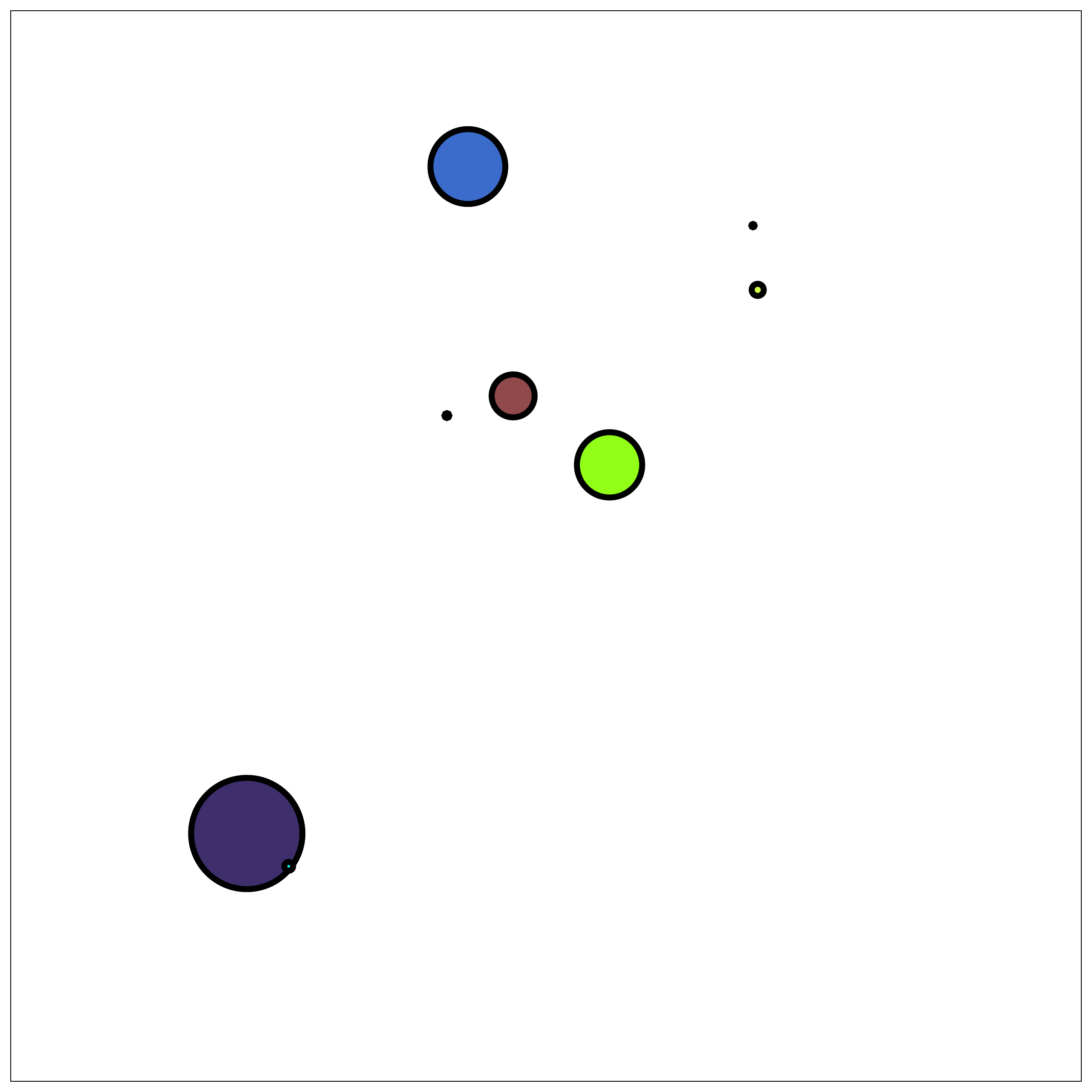}
            \includegraphics[width=0.135\textwidth]{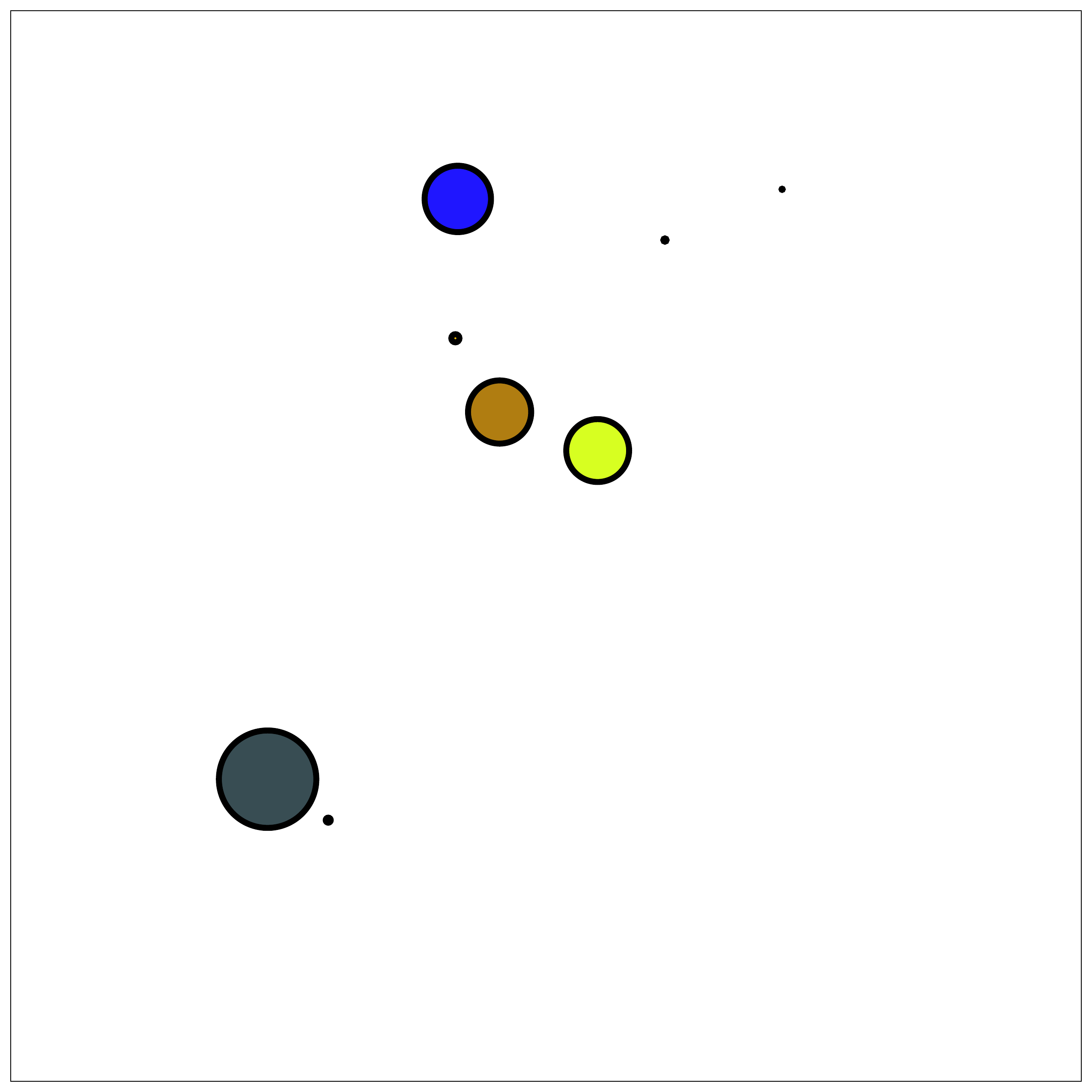}
            \includegraphics[width=0.135\textwidth]{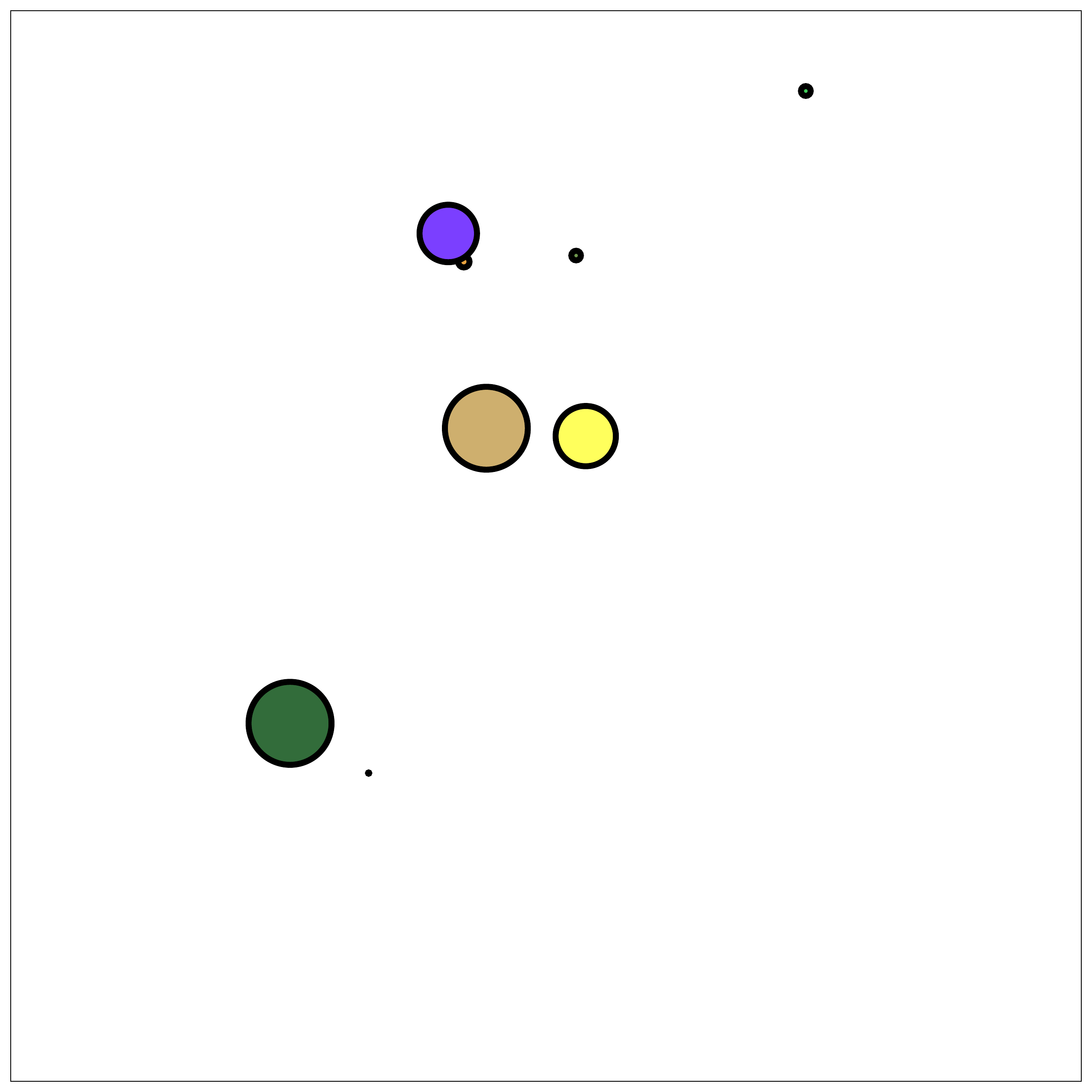}
            \includegraphics[width=0.135\textwidth]{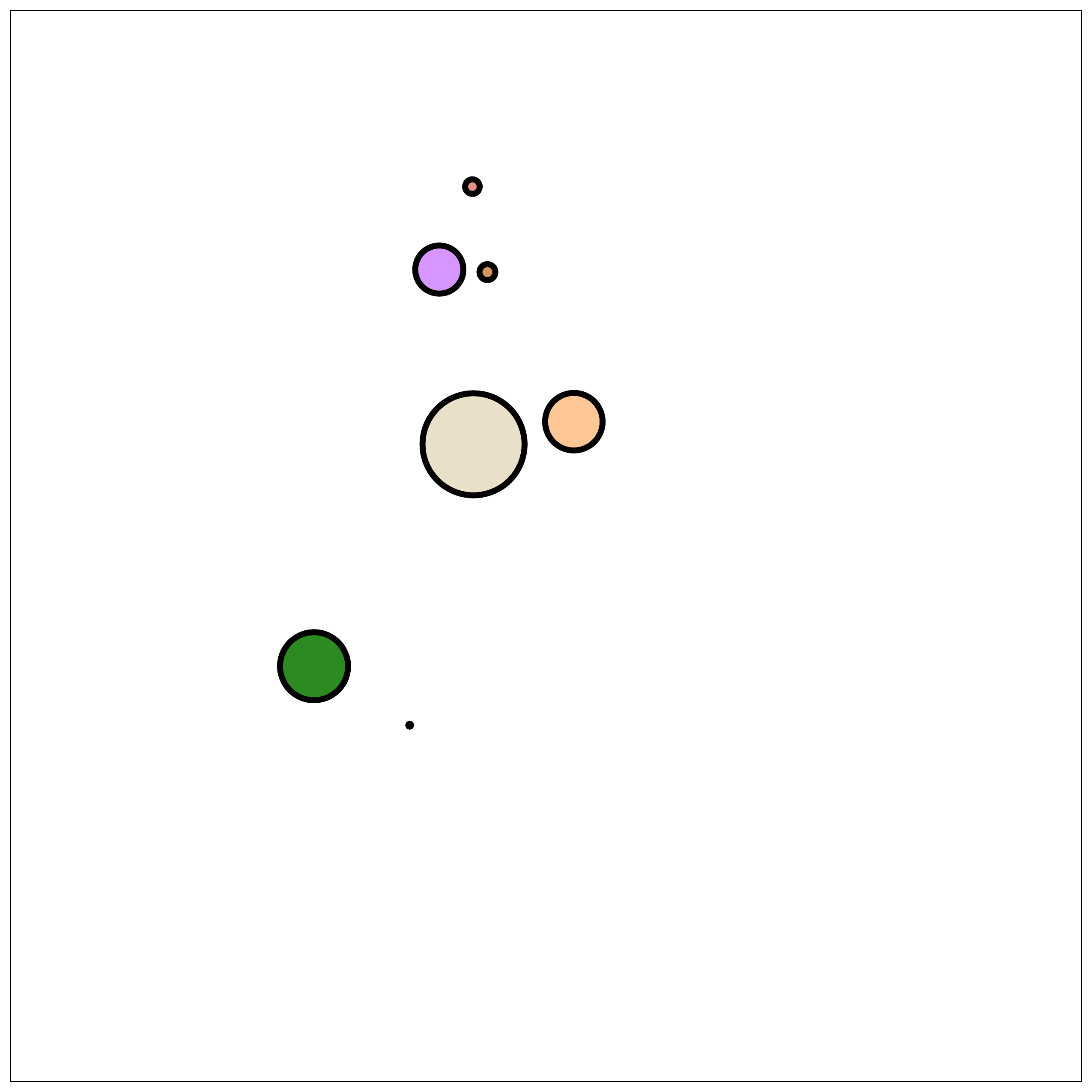}
            \includegraphics[width=0.135\textwidth]{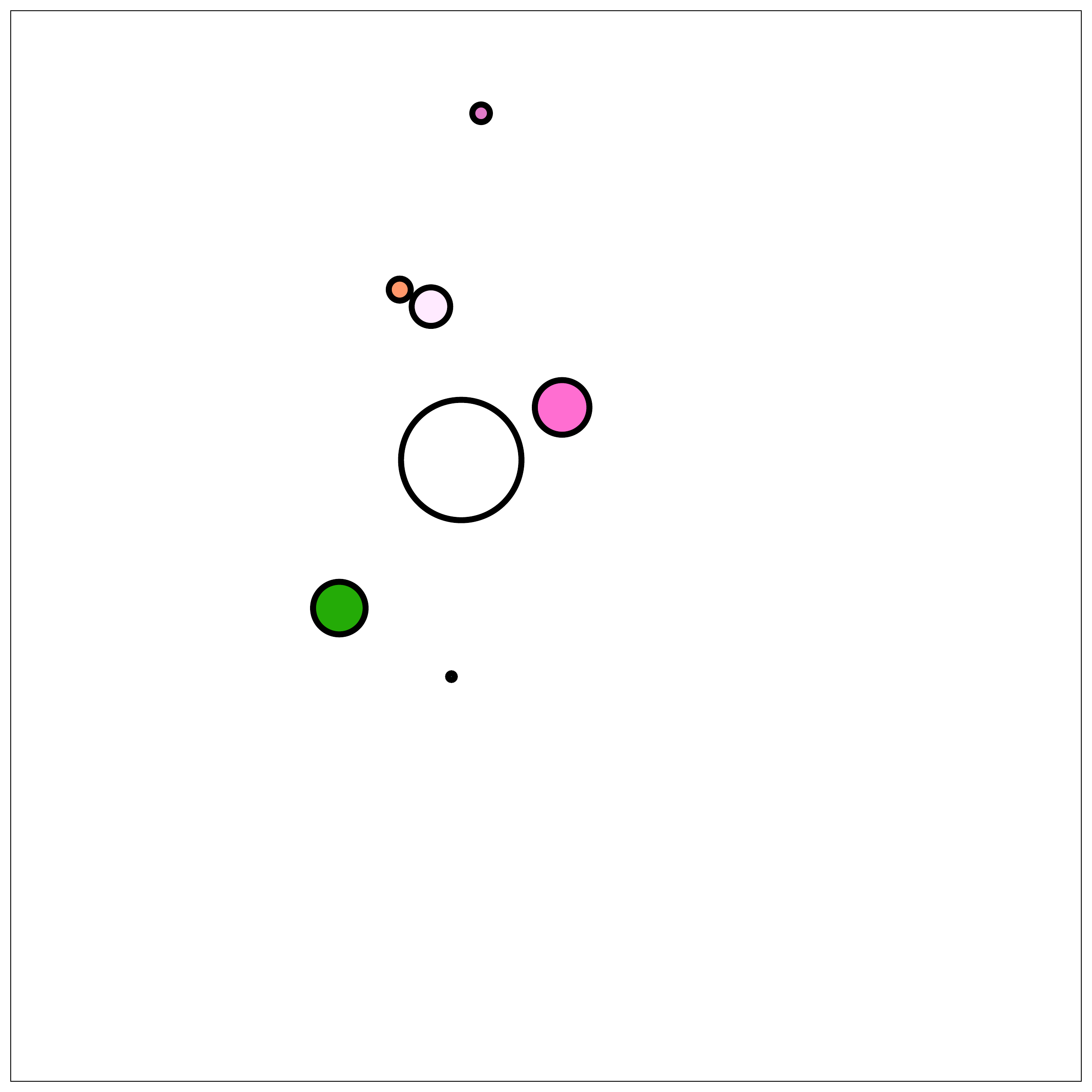}
            \includegraphics[width=0.135\textwidth]{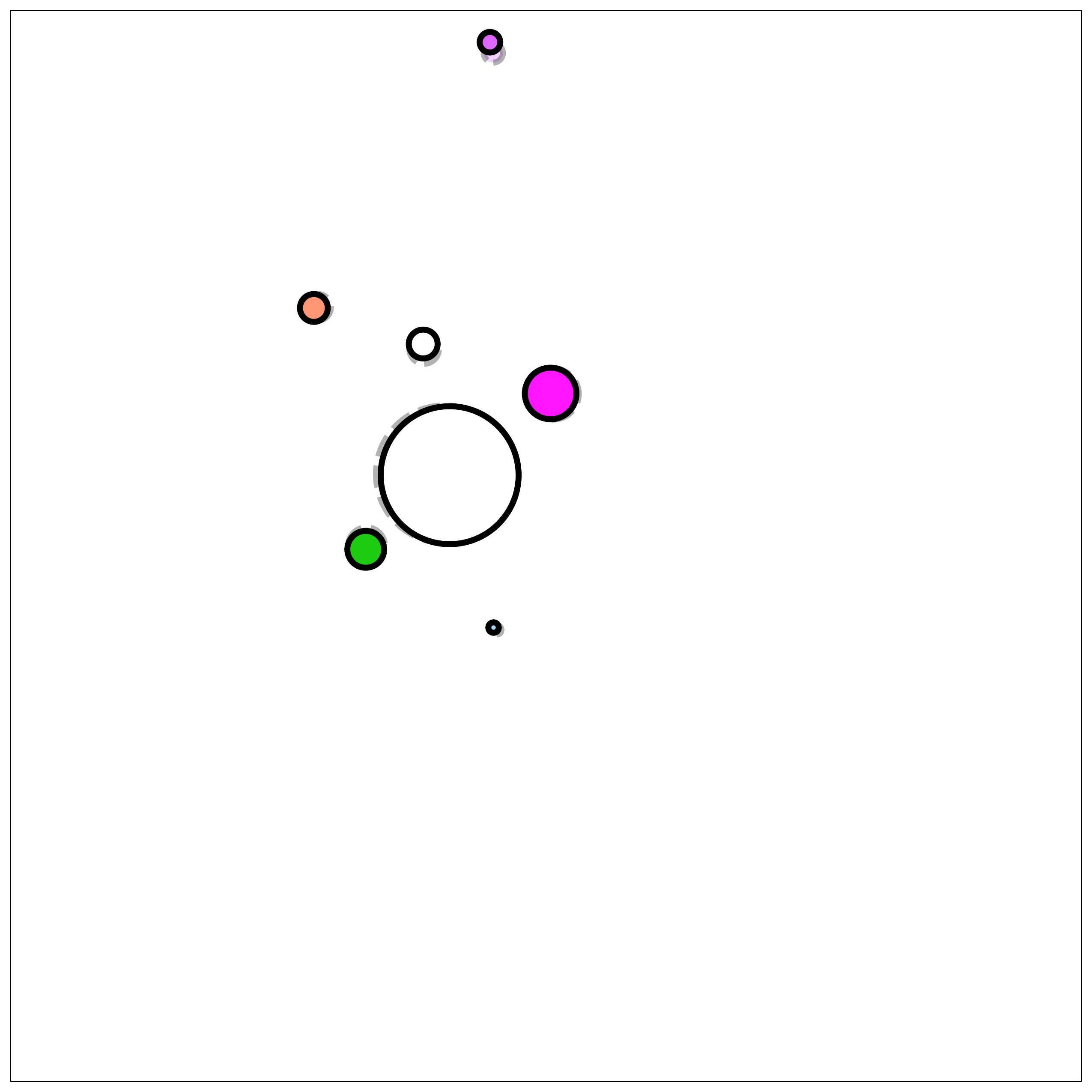}
            \caption{PISA}
    \end{subfigure}
    \begin{subfigure}[b]{.95\textwidth}
            \centering
            \includegraphics[width=0.135\textwidth]{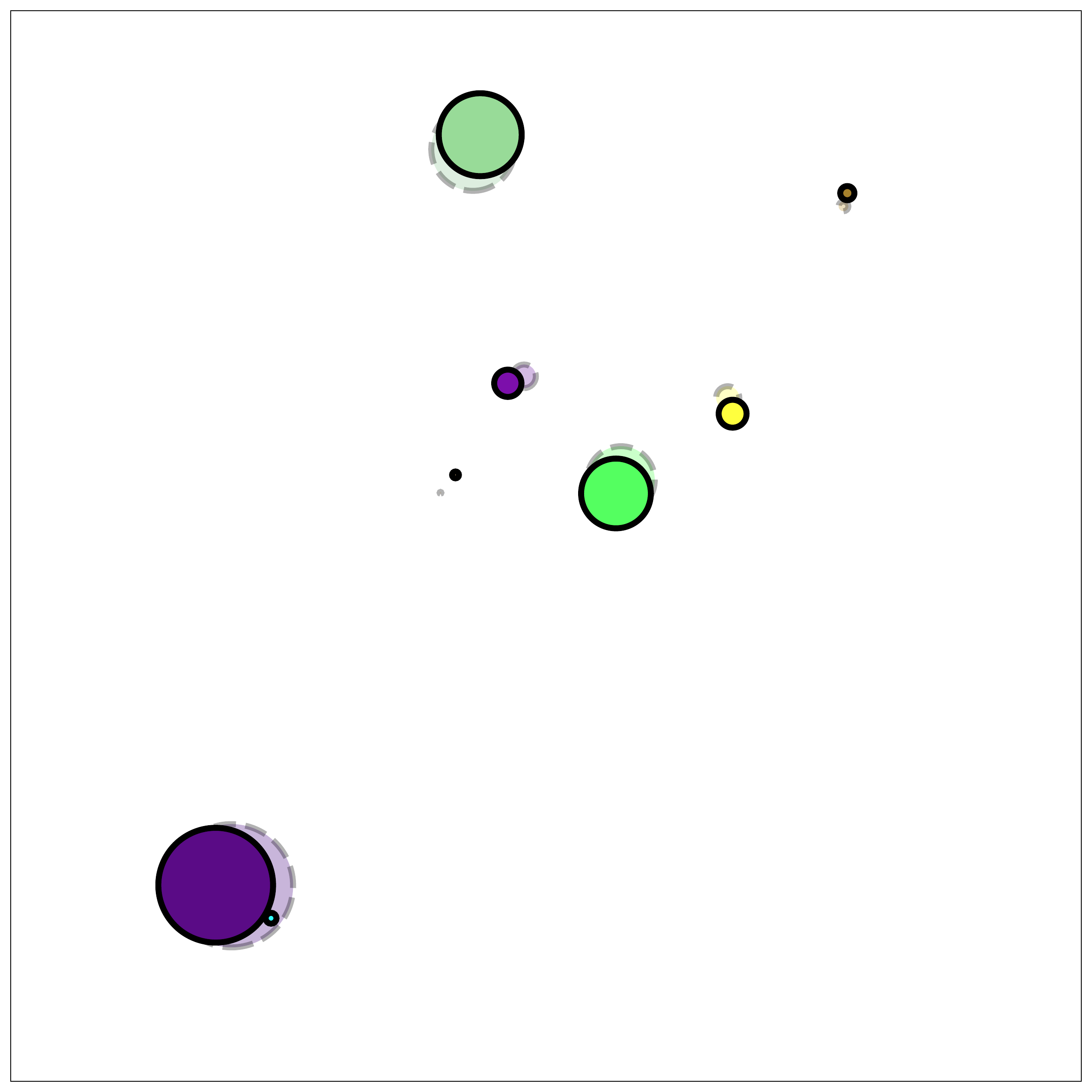}
            \includegraphics[width=0.135\textwidth]{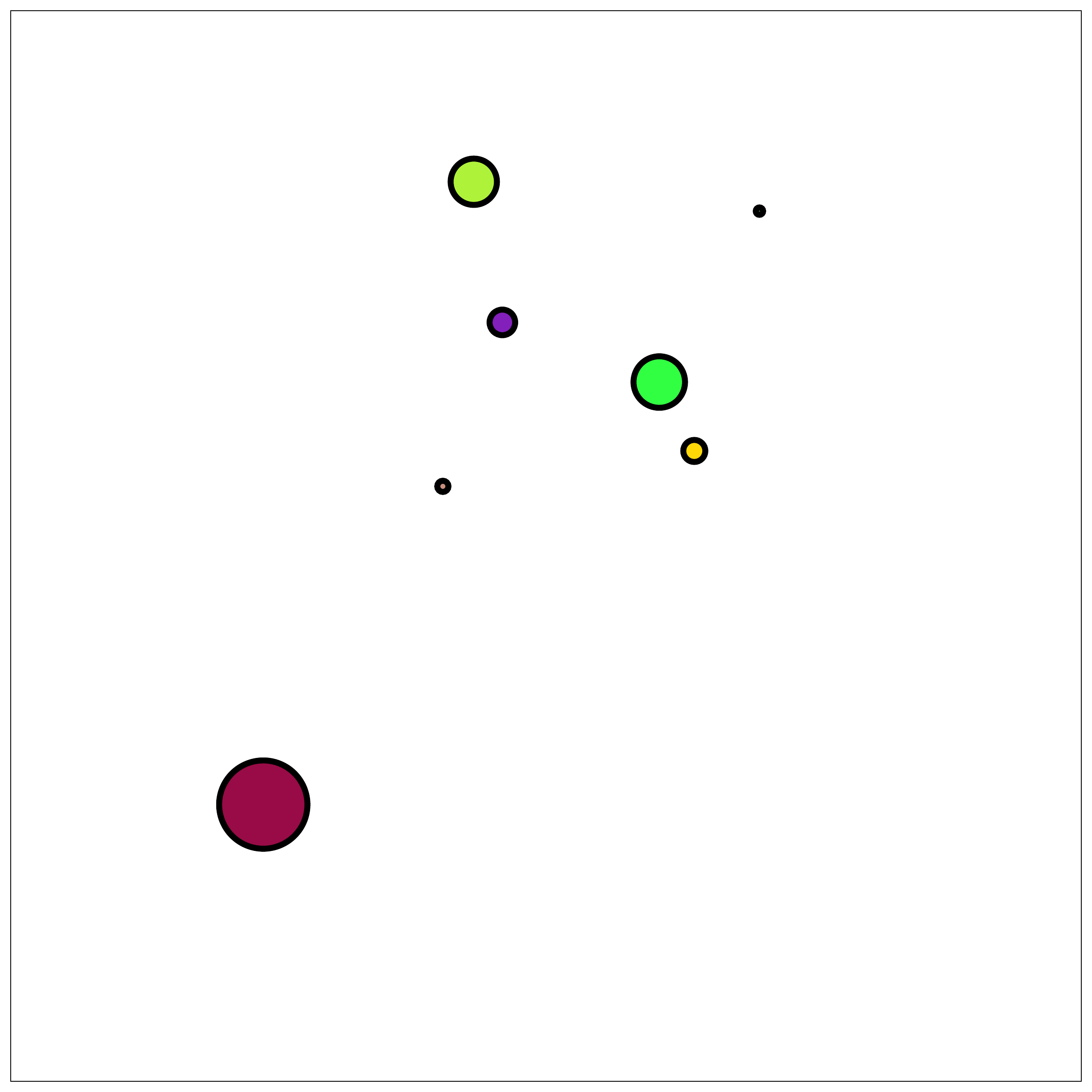}
            \includegraphics[width=0.135\textwidth]{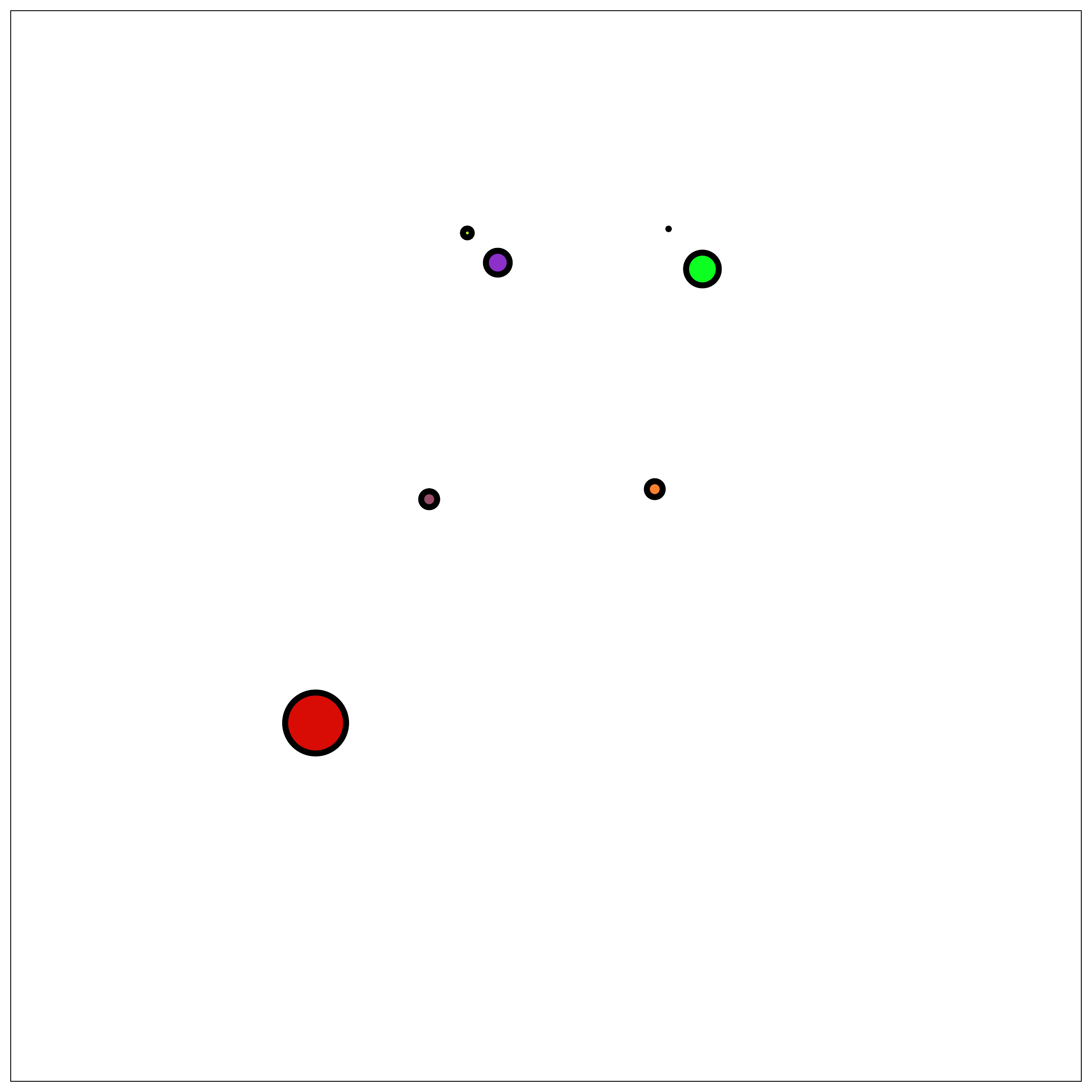}
            \includegraphics[width=0.135\textwidth]{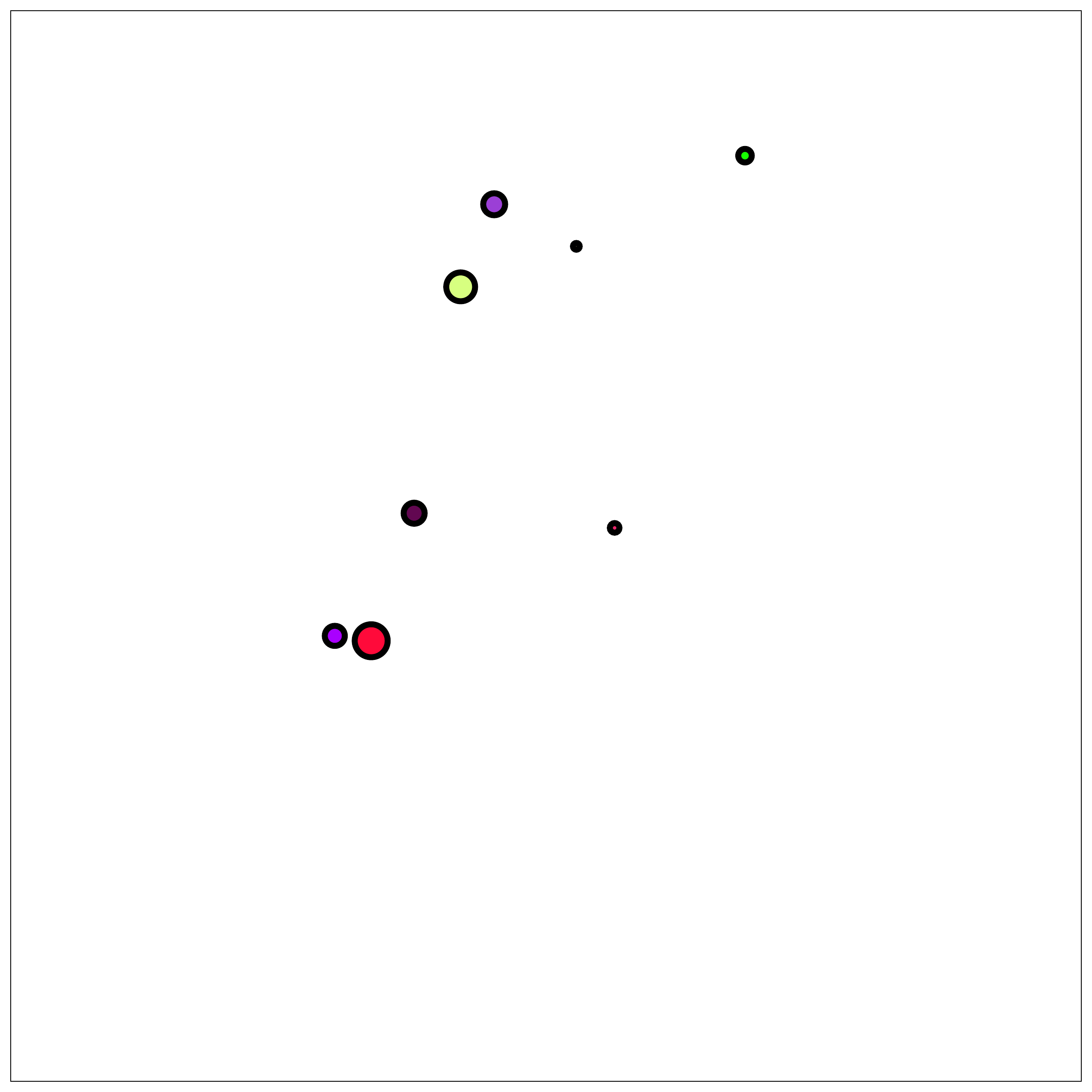}
            \includegraphics[width=0.135\textwidth]{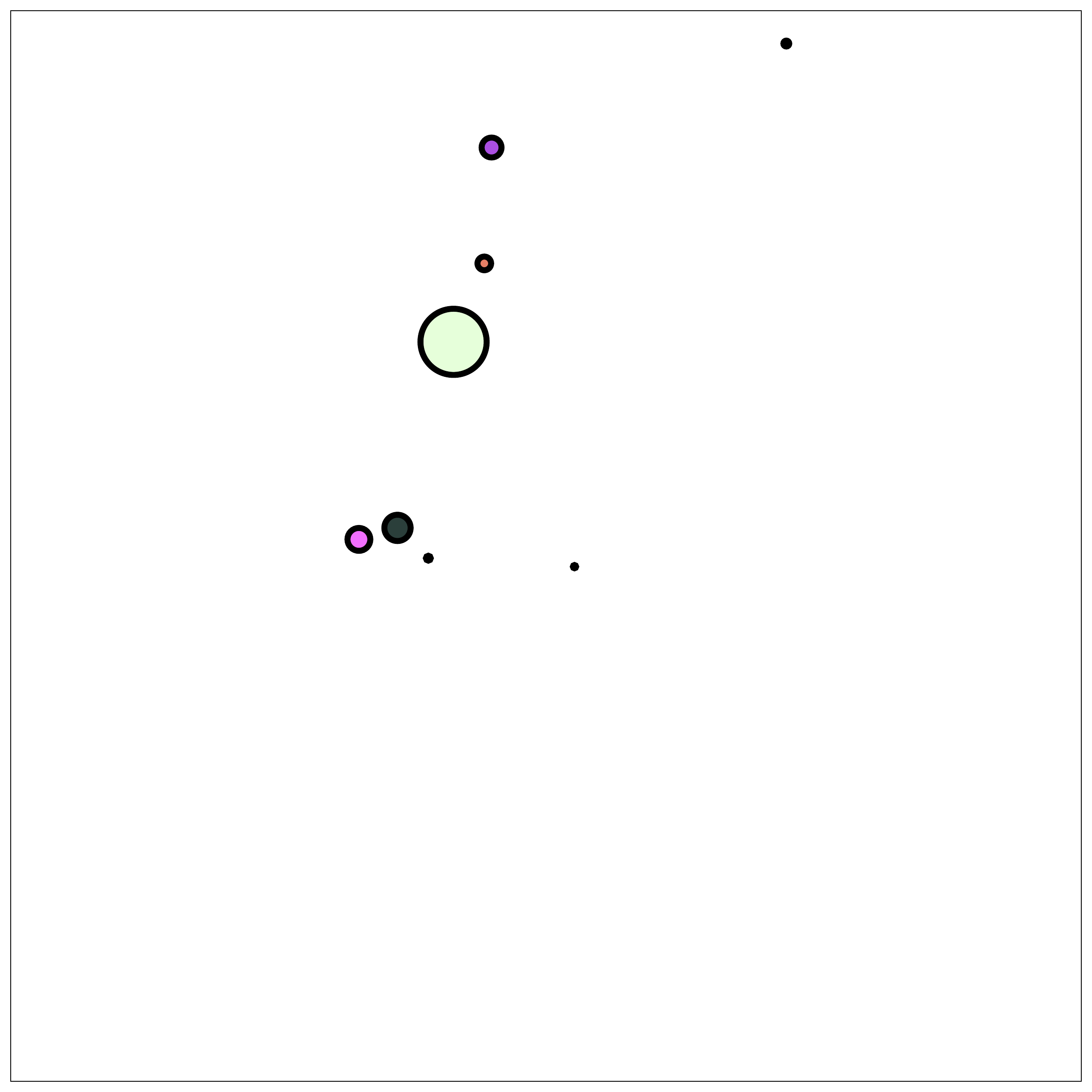}
            \includegraphics[width=0.135\textwidth]{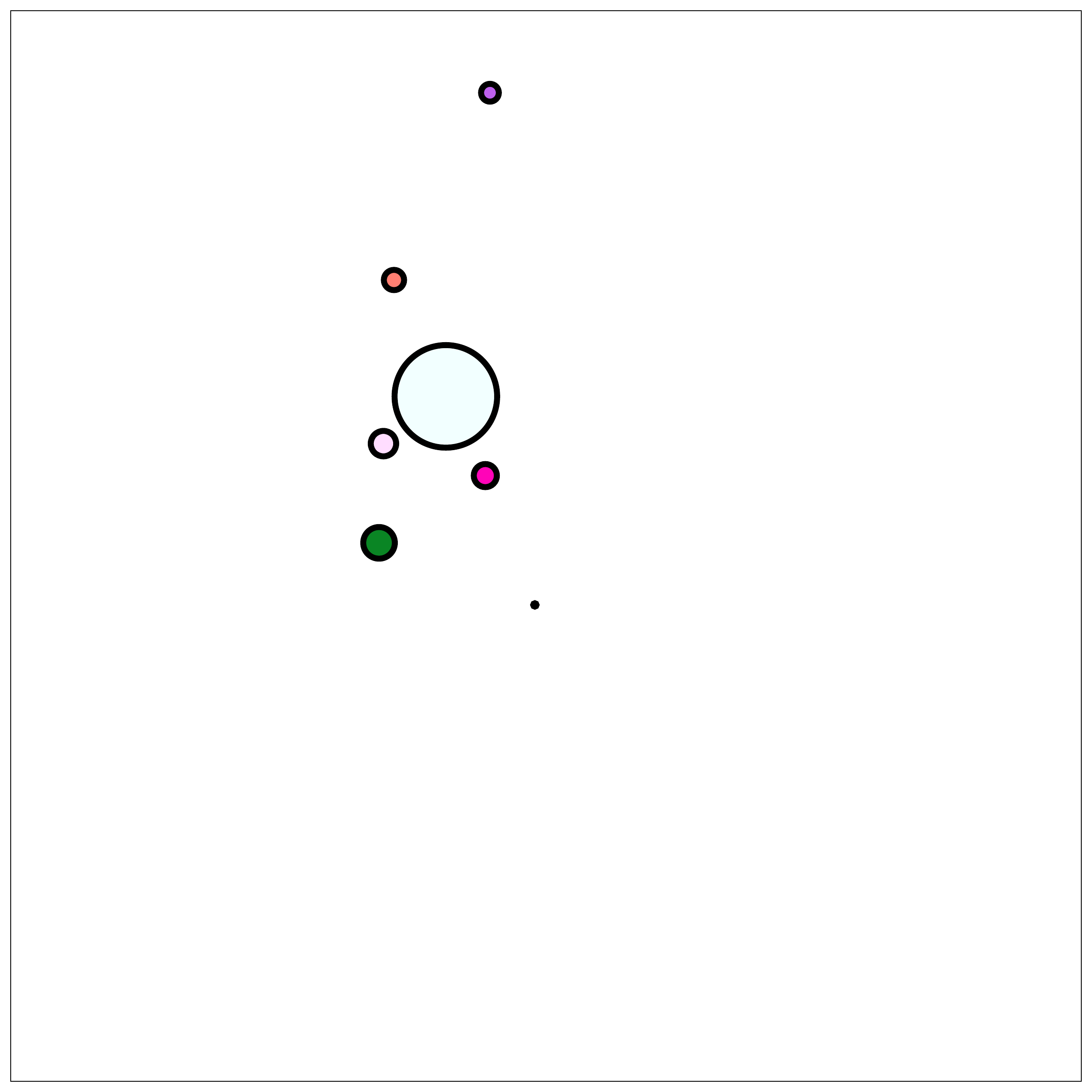}
            \includegraphics[width=0.135\textwidth]{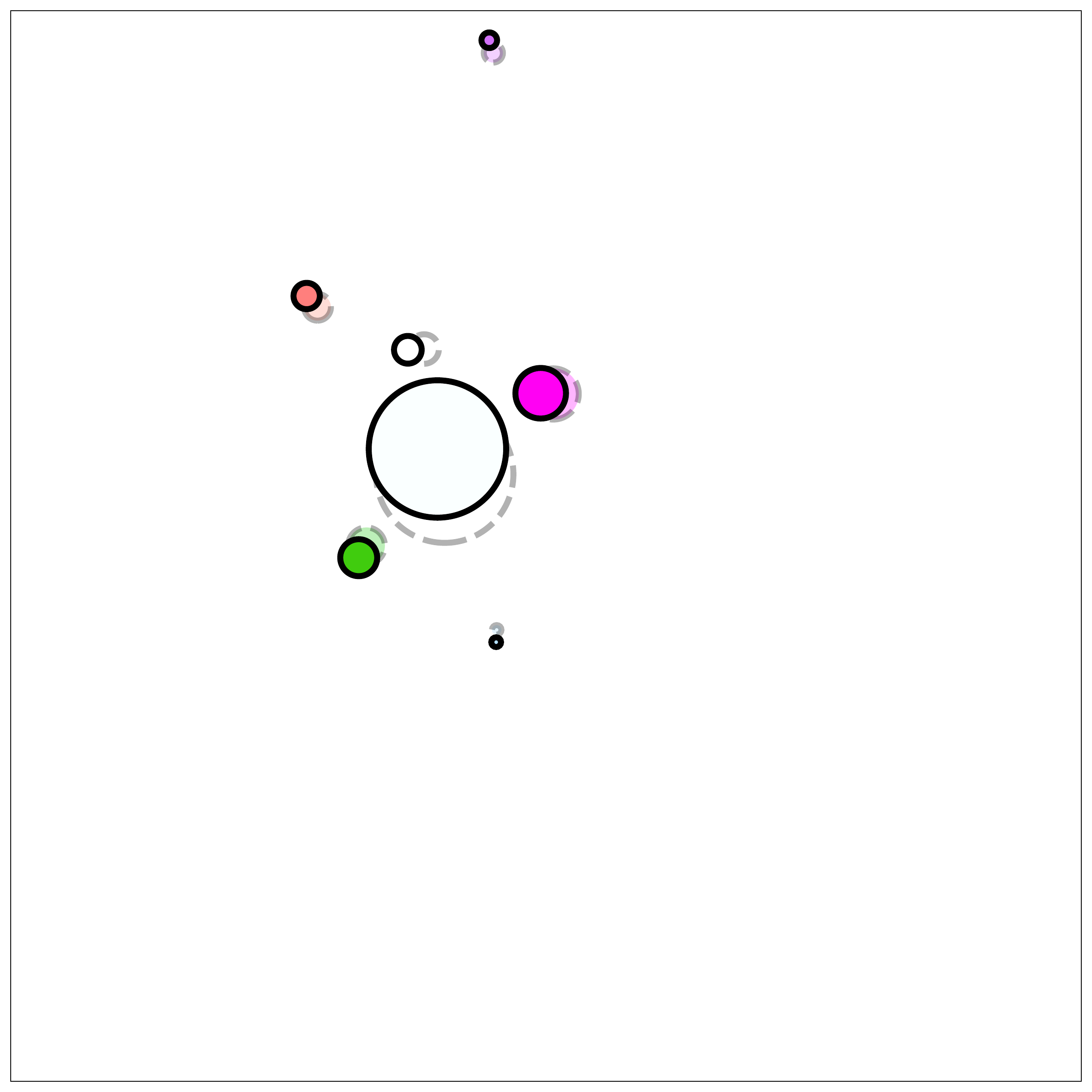}
            \caption{GRU}
    \end{subfigure}
    \begin{subfigure}[b]{.95\textwidth}
            \centering
            \includegraphics[width=0.135\textwidth]{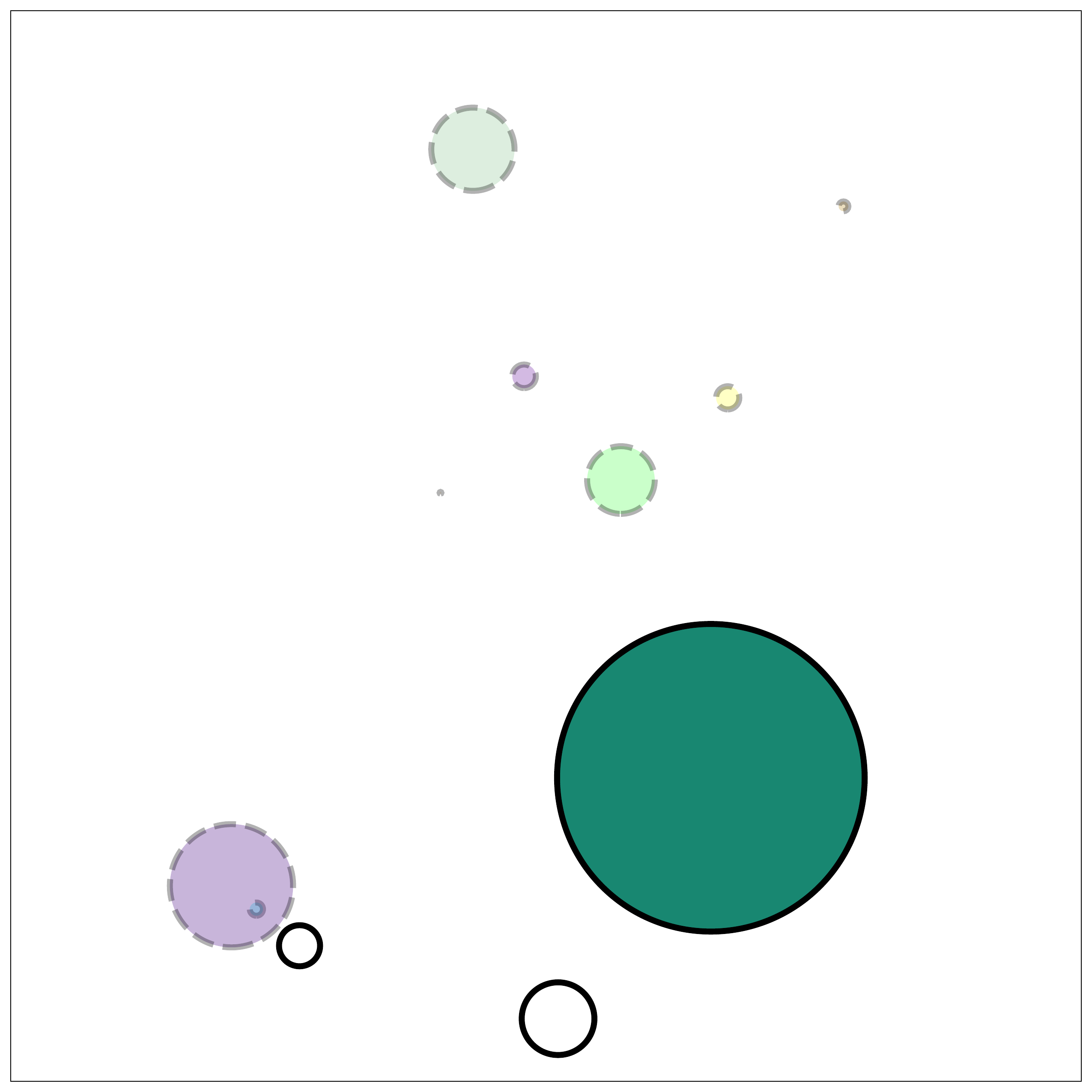}
            \includegraphics[width=0.135\textwidth]{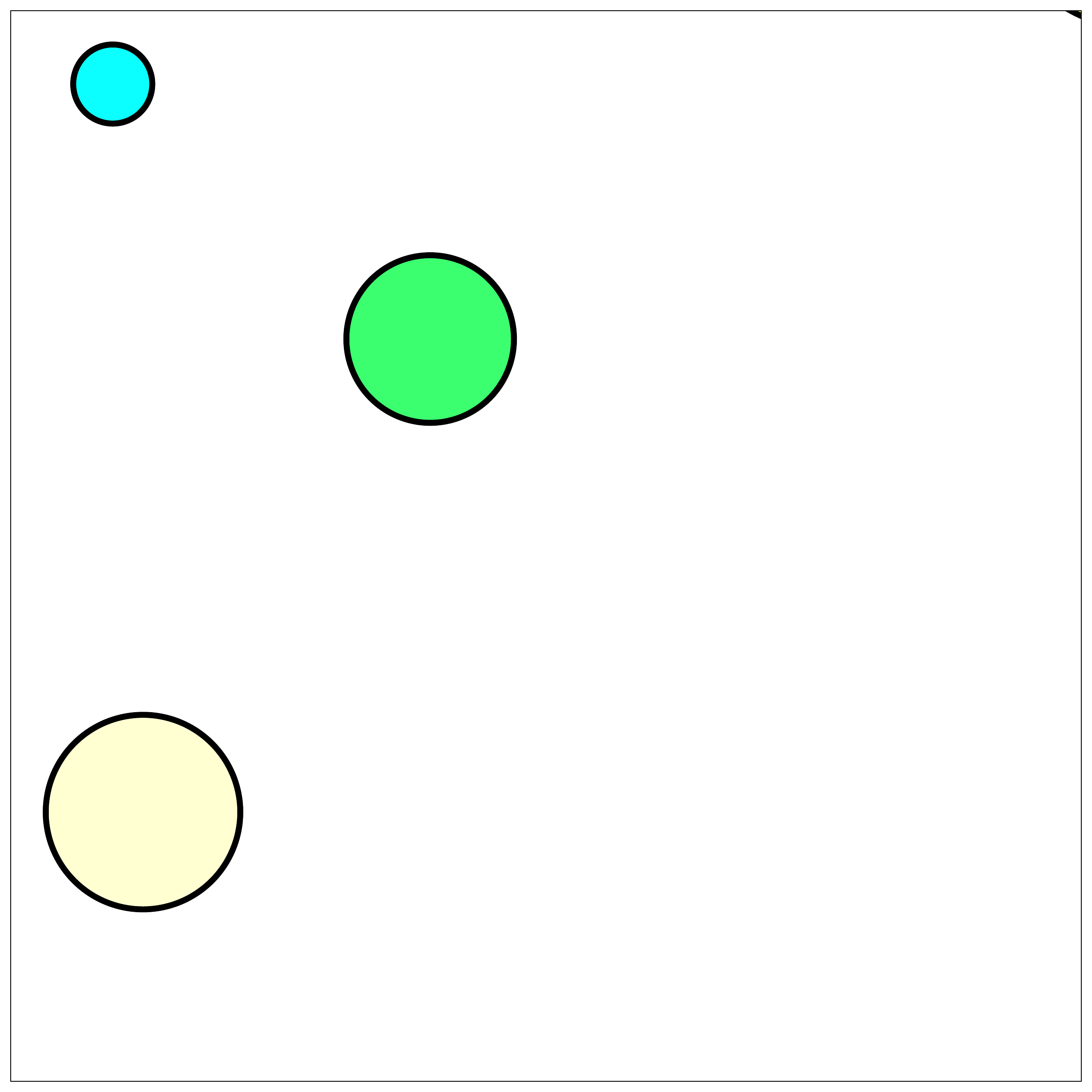}
            \includegraphics[width=0.135\textwidth]{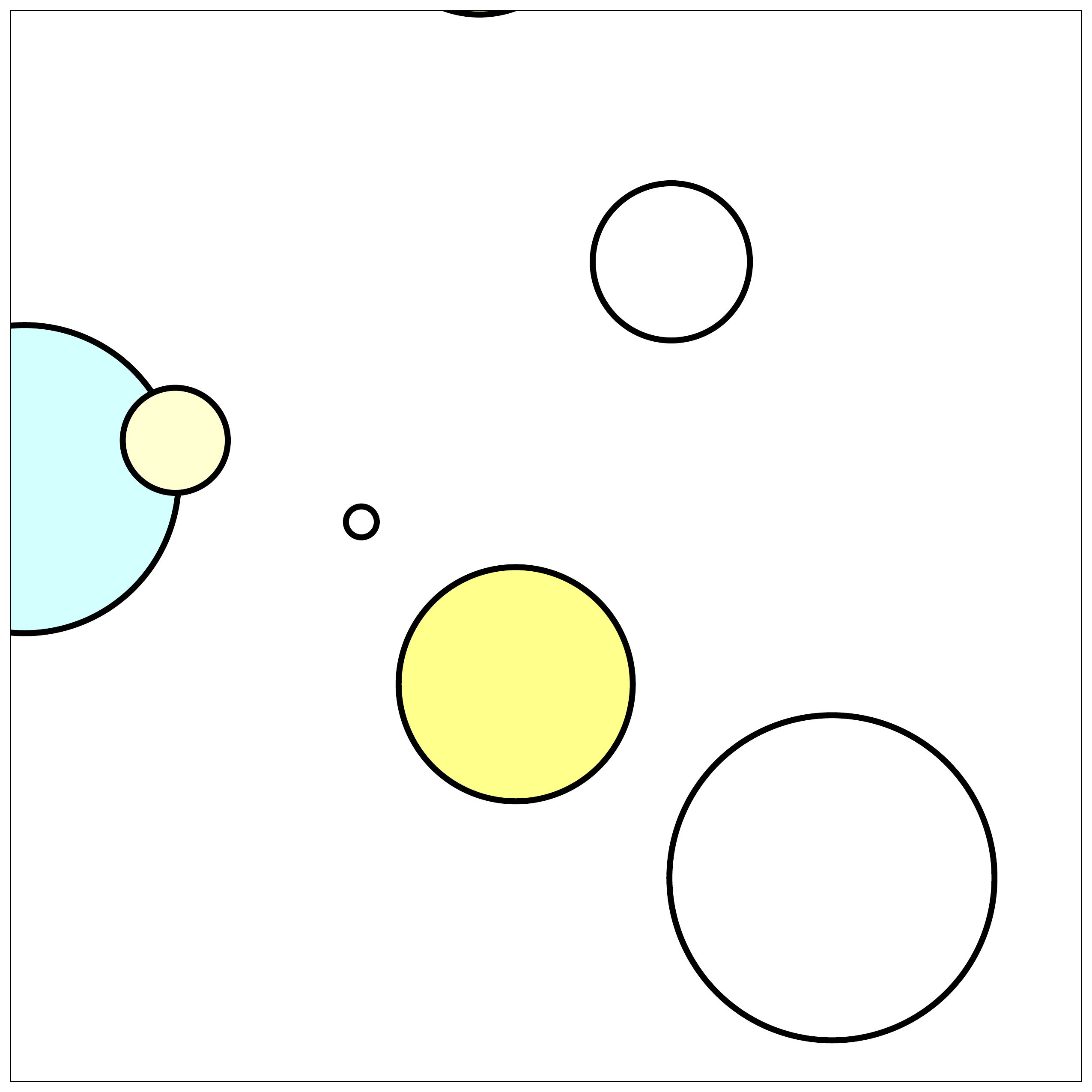}
            \includegraphics[width=0.135\textwidth]{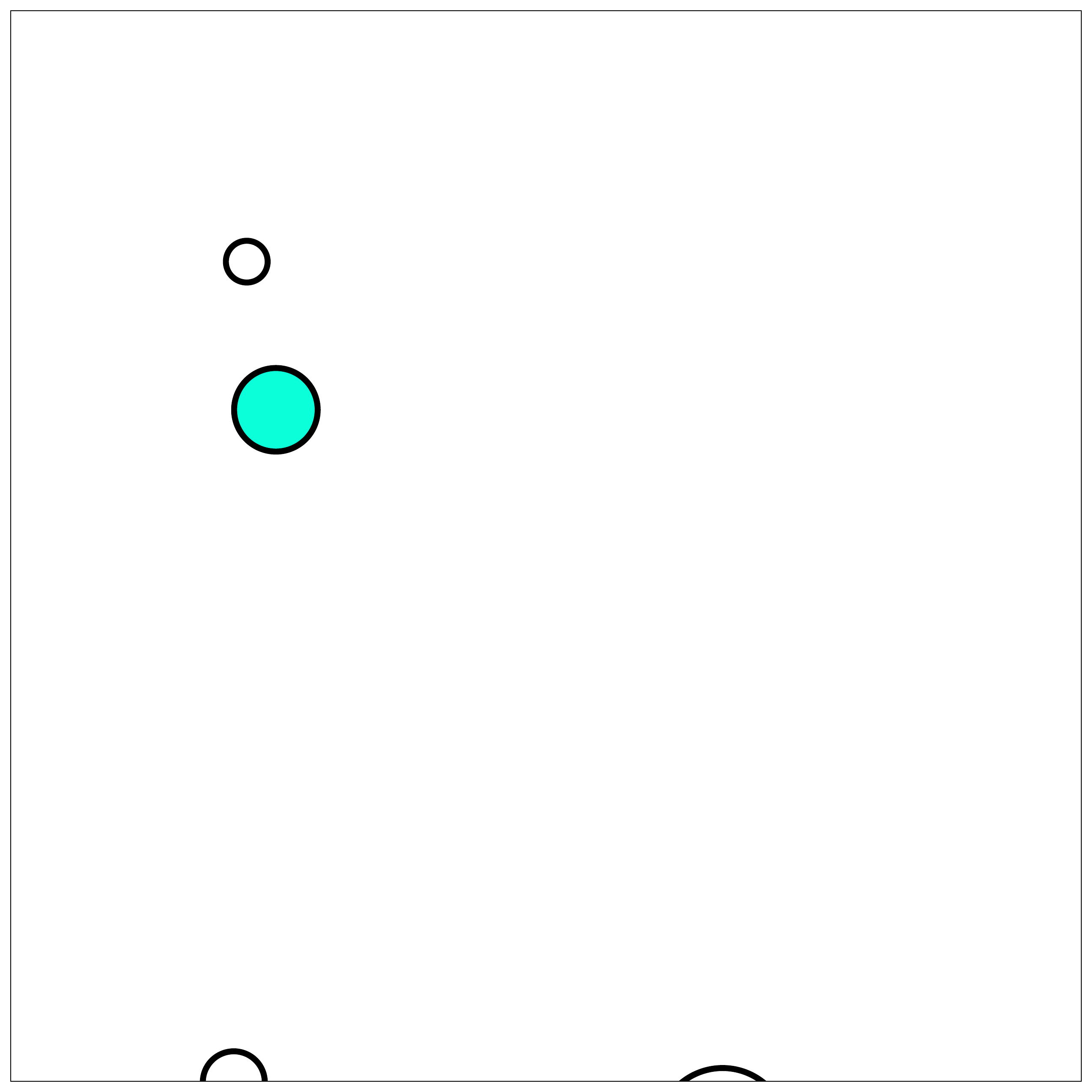}
            \includegraphics[width=0.135\textwidth]{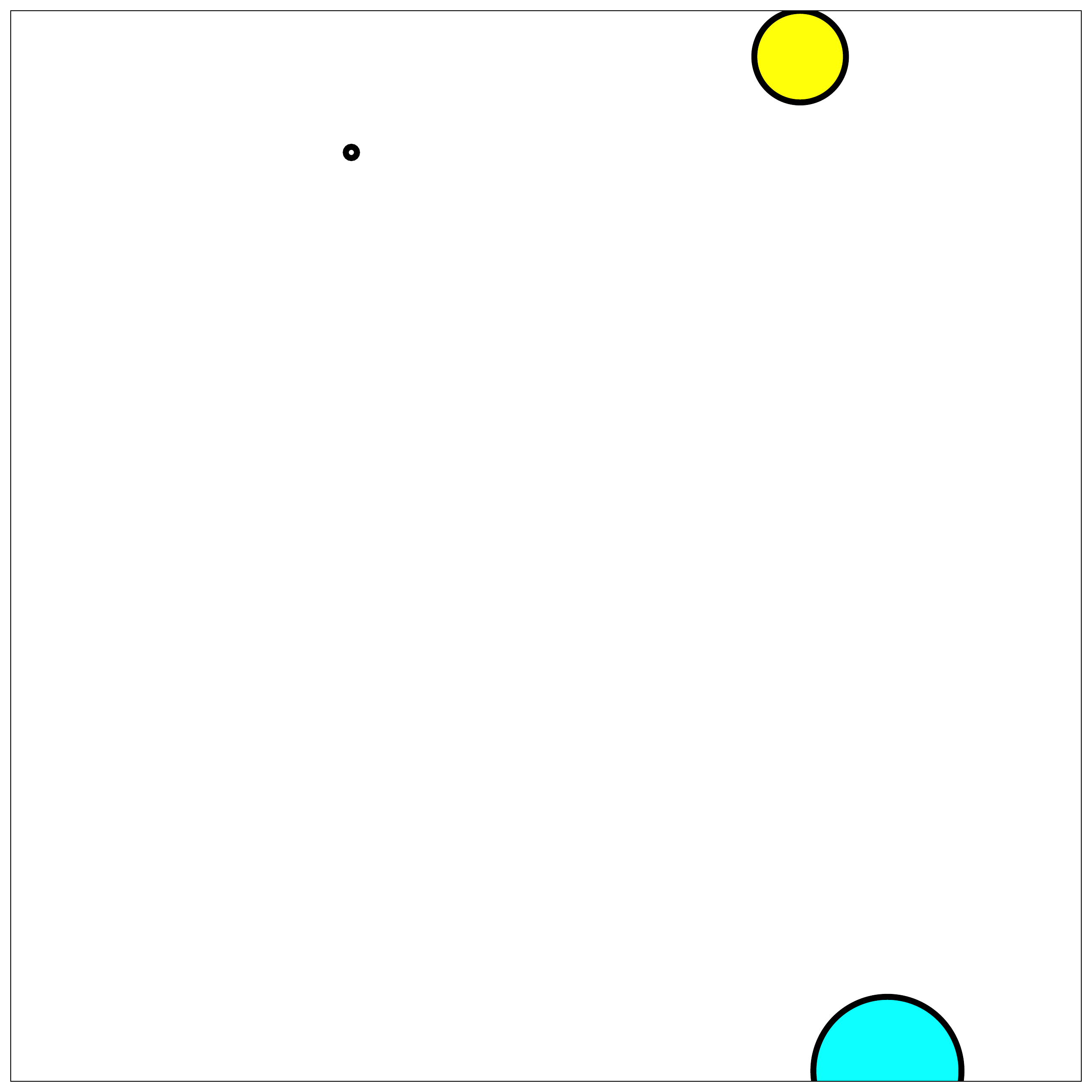}
            \includegraphics[width=0.135\textwidth]{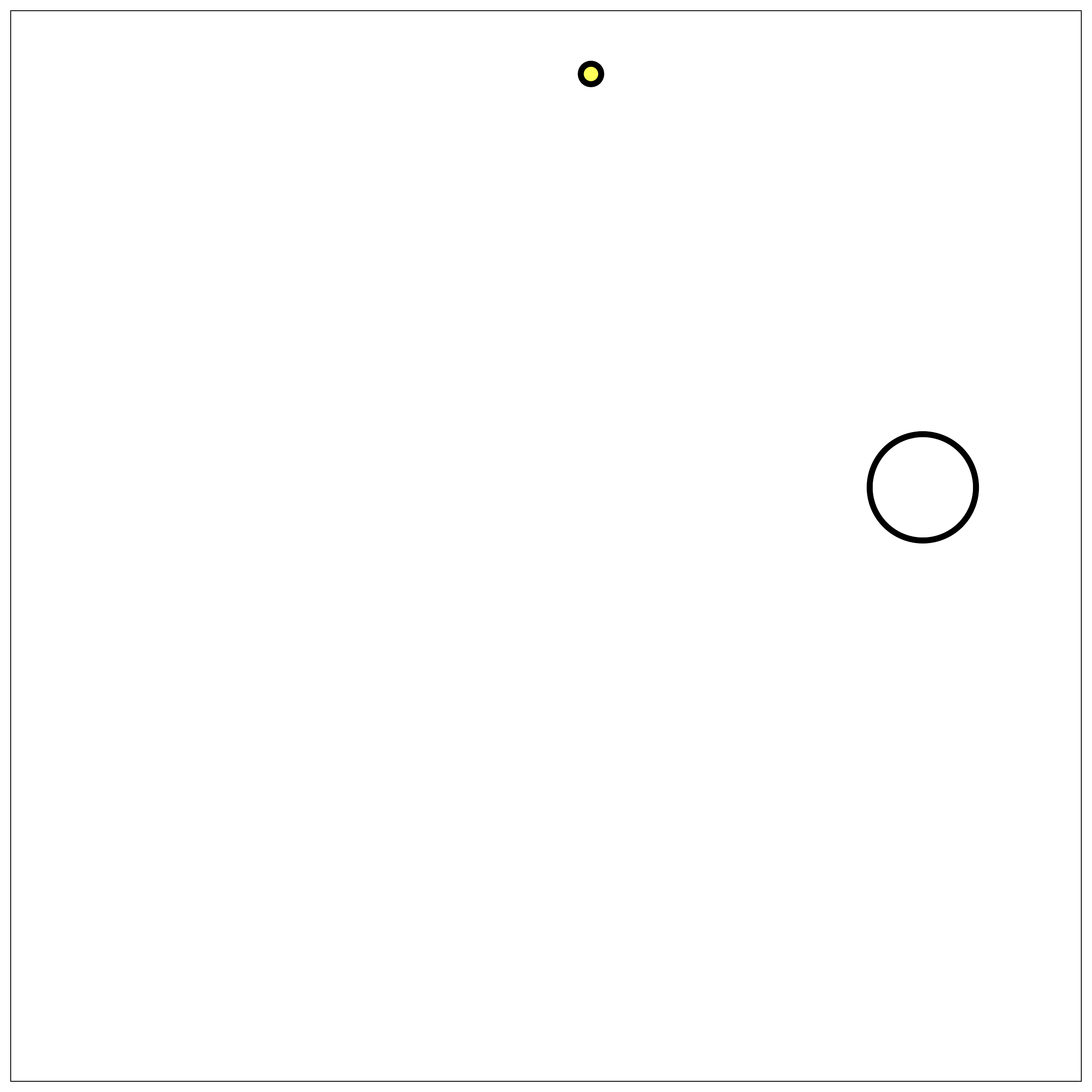}
            \includegraphics[width=0.135\textwidth]{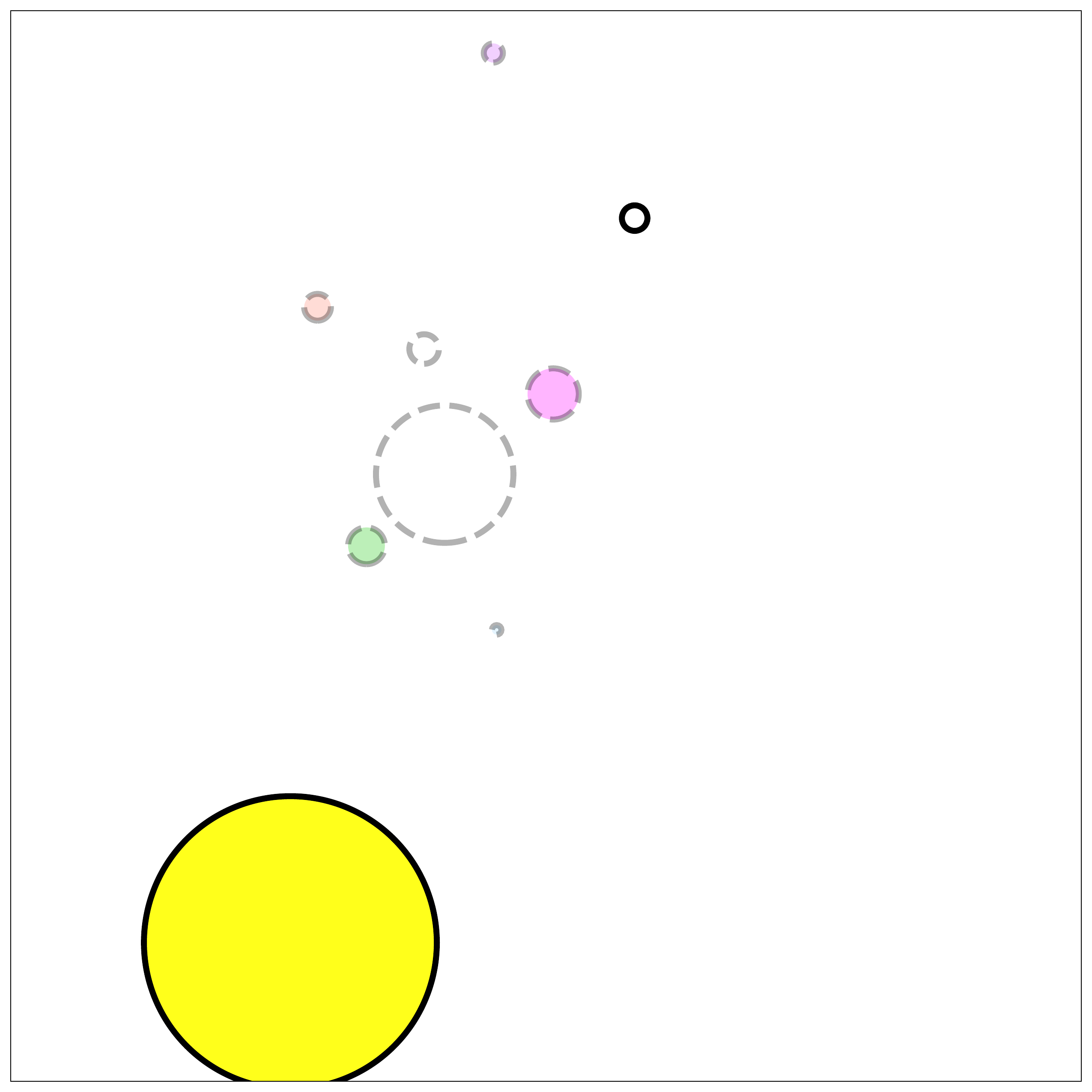}
            \caption{DSPN}
    \end{subfigure}
    \begin{subfigure}[b]{.95\textwidth}
            \centering
            \includegraphics[width=0.135\textwidth]{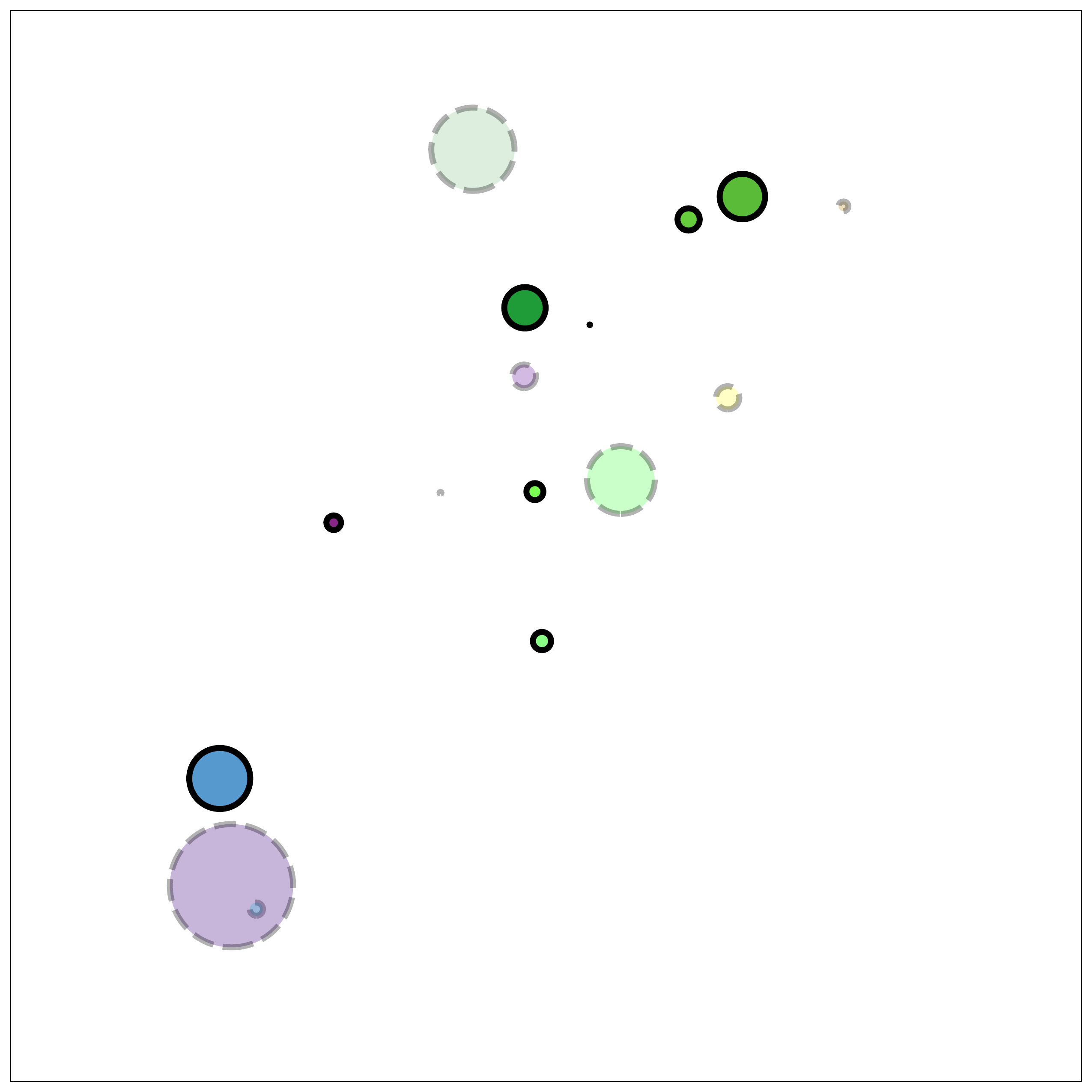}
            \includegraphics[width=0.135\textwidth]{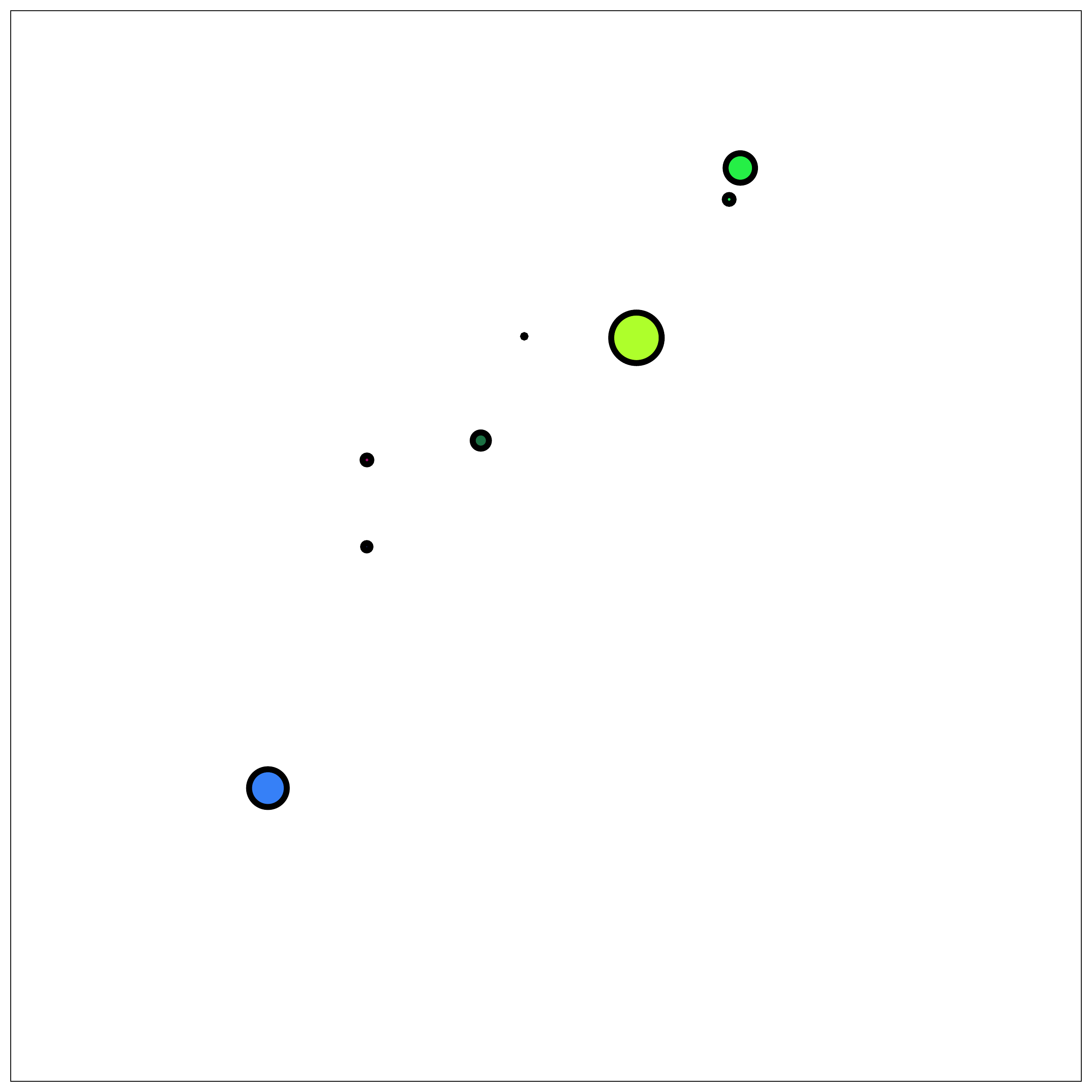}
            \includegraphics[width=0.135\textwidth]{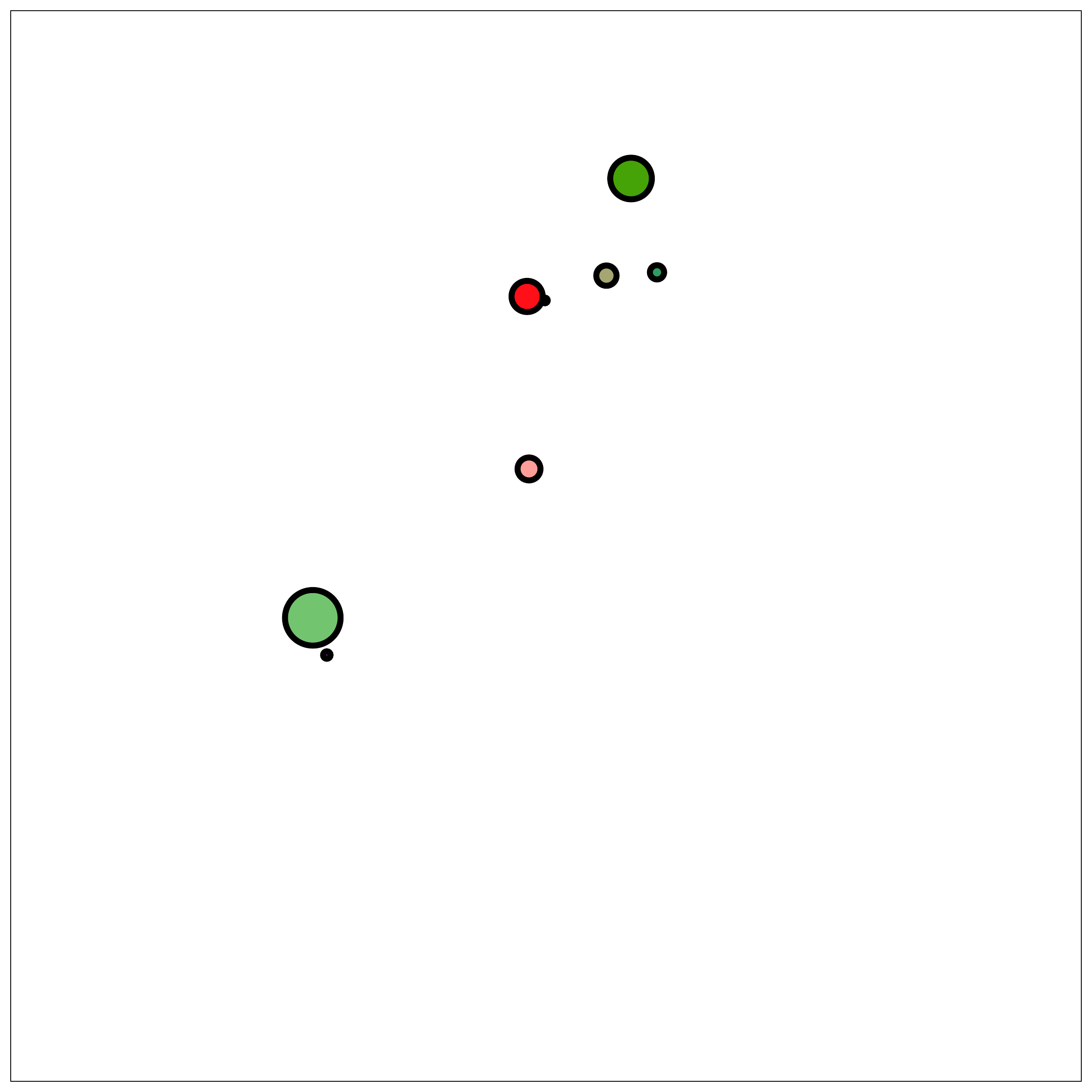}
            \includegraphics[width=0.135\textwidth]{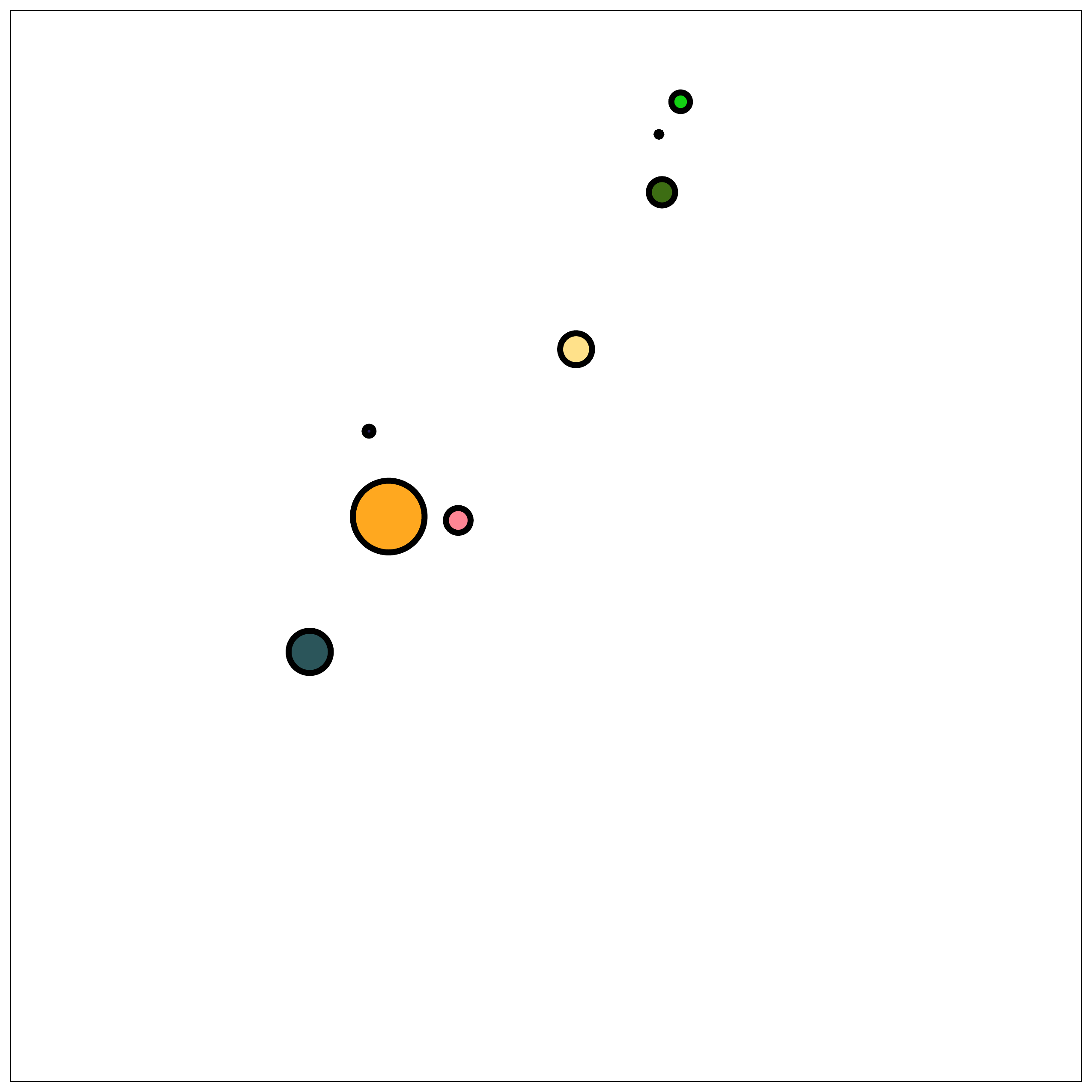}
            \includegraphics[width=0.135\textwidth]{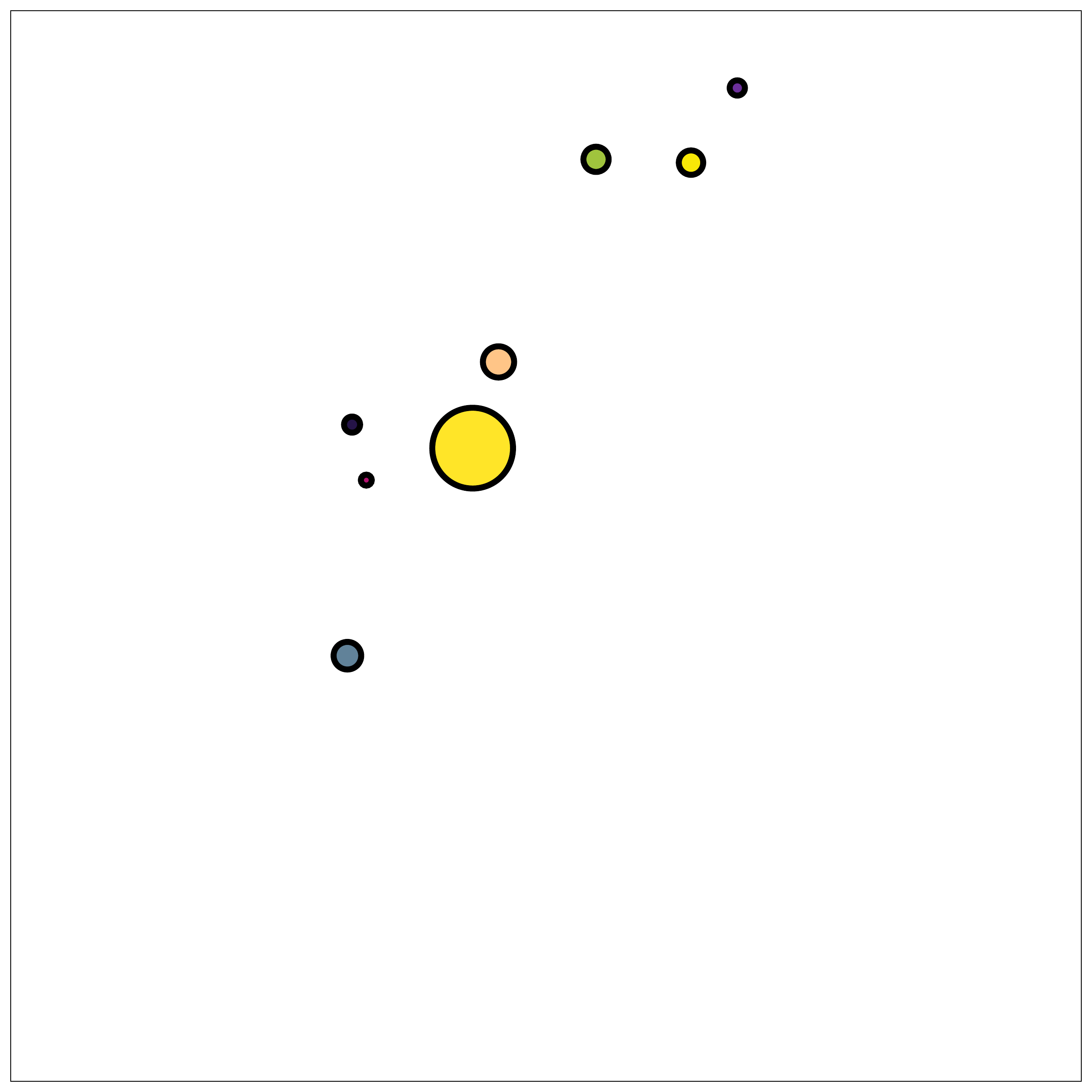}
            \includegraphics[width=0.135\textwidth]{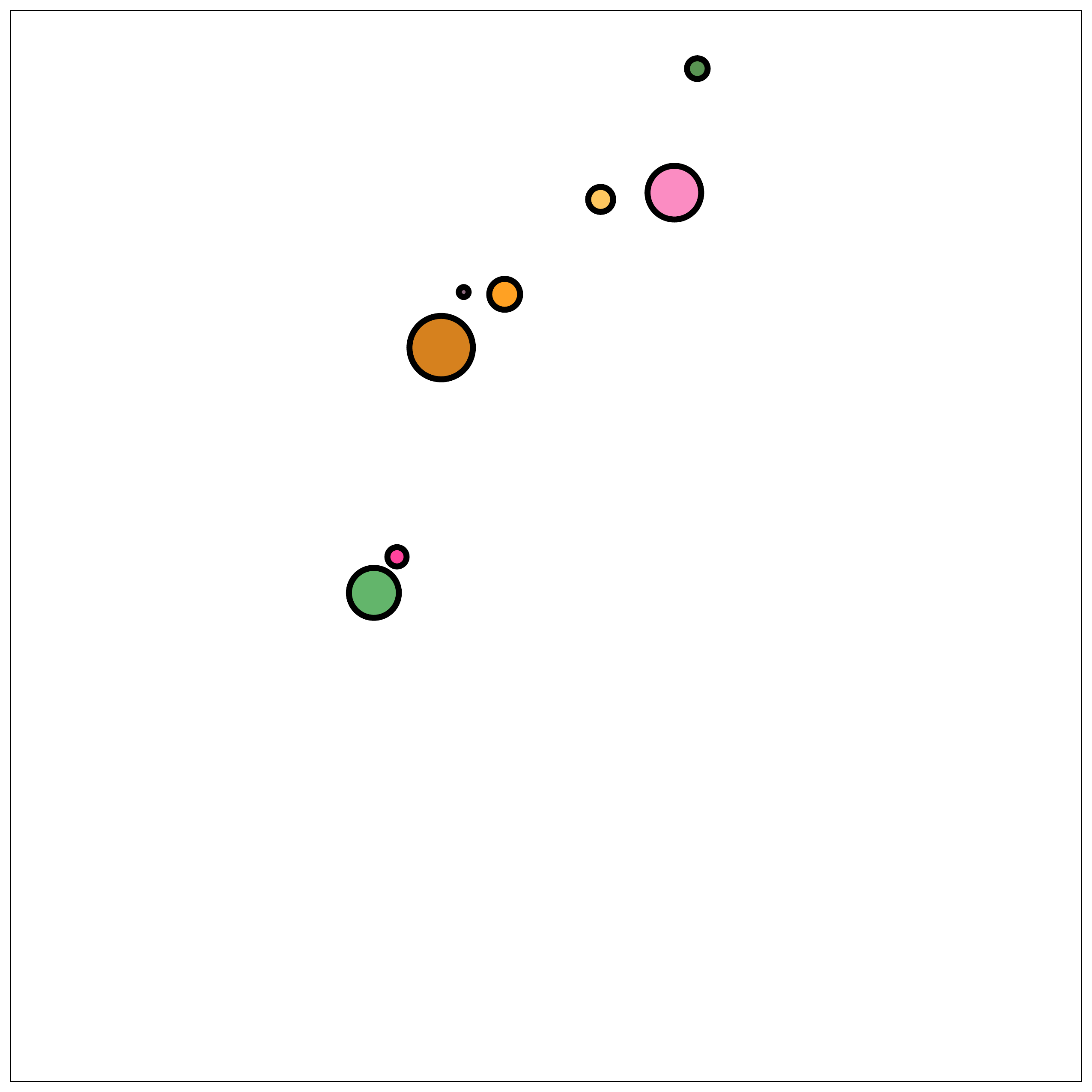}
            \includegraphics[width=0.135\textwidth]{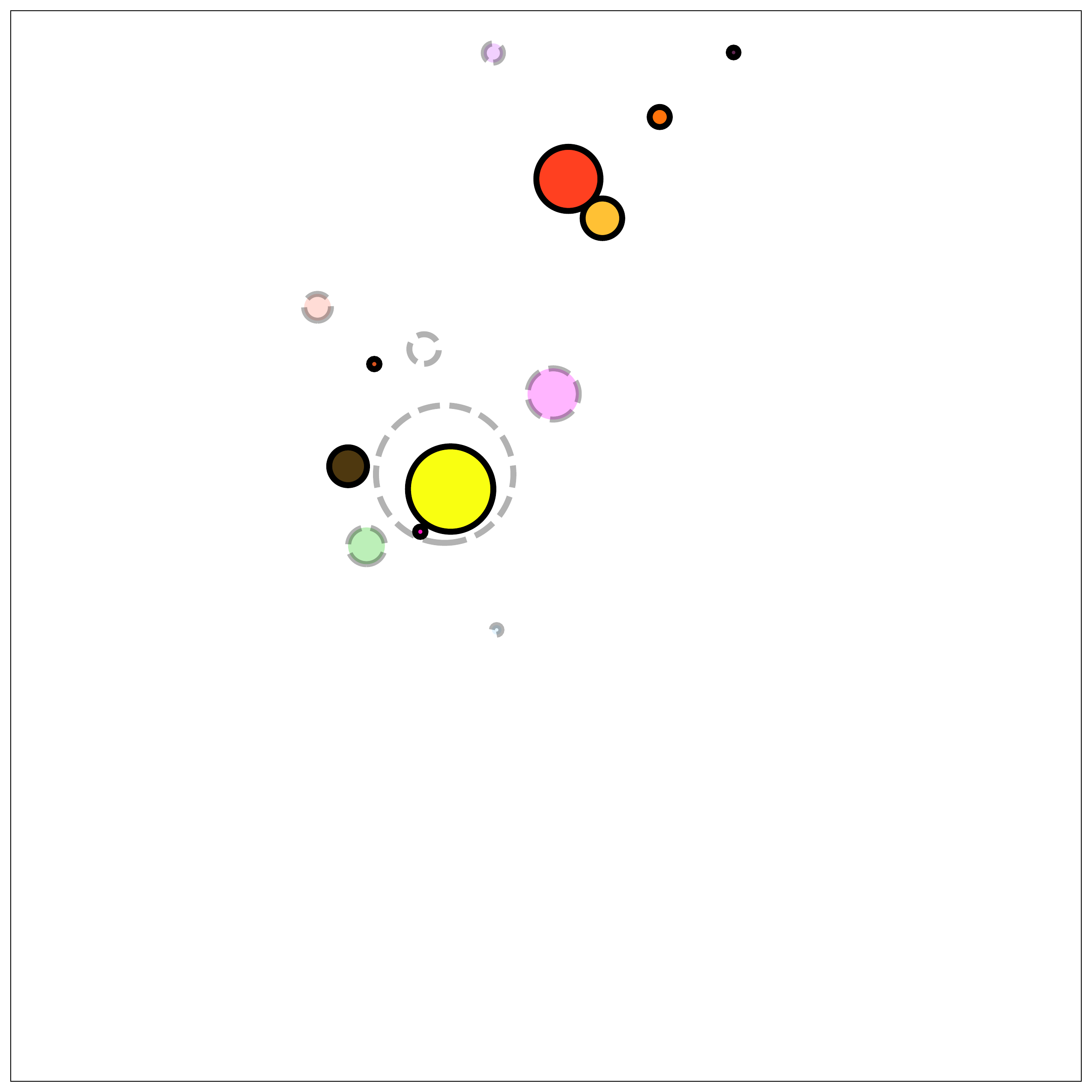}
            \caption{TSPN}
    \end{subfigure}
    \caption{The sensitivity of the latent space for each method. In this experiment, we encode two sets with $8$ elements into representations $z_0$ and $z_1$. Then, we continuously interpolate in the latent space between $z_0$ and $z_1$, and decode the result. In the figure, each panel is calculated as $\phi((1-\alpha) \cdot z_0 + \alpha \cdot z_1)$, from $\alpha=0$ on the left to $\alpha=1$ on the right. All of the methods are evaluated on the same sets $\mathcal{X}_0$ and $\mathcal{X}_1$, which are shown in the $\alpha=0$ and $\alpha=1$ panels with lower opacity.}
    \label{fig:interpolation}
\end{figure*}

\subsection{Encoder}

In the encoder architecture, keys and values are produced by learned networks $\psi_\mathrm{key}$ and $\psi_\mathrm{val}$ (Fig. \ref{fig:sae architecture}). The embedding is generated by taking the element-wise product between each encoded key-value pair, and summing over all elements:
\begin{equation}
    z = \sum_{i=1}^n \left[ \psi_\mathrm{key}(\rho(x_i)) \odot \psi_\mathrm{val}(x_i) \right] + \lambda_\mathrm{enc}(n).
    \label{eq:encoder}
\end{equation}

The input to the key network $\psi_\mathrm{key}$ is generated by a value-to-key mapping function $\rho: \mathbb{R}^{d_x} \mapsto [1..n]$. Although any key can be used to insert any value, the mapping function $\rho$ guarantees permutation invariance by assigning keys to values in a deterministic manner. Since the pairing between keys and values has a negligible effect on the output, it is not necessary to learn $\rho$---it is implemented with a randomly initialised linear layer. In practice, we also transform the output of $\rho$ to the onehot space before passing it to $\psi_\mathrm{key}$ in order to remove the inherent ordering in the keys.

In addition to encoding the elements from the input set, we also encode its cardinality. In equation \eqref{eq:encoder}, $\lambda_\mathrm{enc} : \mathbb{Z}^+_0 \mapsto \mathbb{R}^{d_z}$ encodes the cardinality with a single linear layer. It is important that this mapping is linear, as it maintains additivity within the latent space (see Appendix \ref{appendix: additivity}).

\subsection{Decoder}

The decoder architecture closely mirrors the encoder architecture, extracting values in a similar manner to how they were inserted (Fig. \ref{fig:sae architecture}). The decoder uses $\lambda_\mathrm{dec}$ to predict the number of outputs $\hat{n}$, and then uses $\phi_\mathrm{key}$ to produce queries $q_i$:
\begin{align}
  \hat{n} &= \lambda_\mathrm{dec}(z)\\
  \{ q_{1}, \dots q_{\hat{n}} \} &= \{ \phi_\mathrm{key}(i) \mid i \in [1 .. \hat{n}] \} 
\end{align}
By computing the element-wise product between the latent state and the queries, the decoder produces an element-specific hidden state $z \odot q_i$. Finally, a decoder network $\phi_\mathrm{dec}$ processes each hidden state, producing the reconstructed set elements
\begin{align}
    \hat{\mathcal{X}} = \{\phi_\mathrm{dec}(z \odot q_i) \mid i \in [1..\hat{n}]\}.
\end{align}

\subsection{Training}

\sloppy The full autoencoder is trained in an unsupervised manner with standard MSE reconstruction loss $\mathcal{L}_\mathrm{mse} = \sum_i^n (\hat{x}_i - x_{\rho(x_i)})^2$. In addition, we use a separate loss to train the networks $\lambda_\mathrm{enc}$ and $\lambda_\mathrm{dec}$, which encode and predict the set's cardinality: $\mathcal{L}_\mathrm{size} =(n - \lambda_\mathrm{dec}(z))^2$.

This training scheme is much more efficient than that of algorithms like DSPN \cite{dspn} and TSPN \cite{tspn}, which use Hungarian loss ($O(n^3)$ complexity) for smaller sets and Chamfer loss ($O(n^2)$ complexity) for larger sets. The latter requires solving an optimisation problem at each step: $\mathcal{L}_\mathrm{hung} = \min_P ||PX - \hat{X}||^2$, where $X, \hat{X} \in \mathbb{R}^{n \times d}$ are matrices of stacked elements from the ground truth and predicted sets, $P \in \{0,1\}^{n \times n}$ is a permutation matrix, and the magnitude $||\cdot||$ represents the row-wise Euclidean norm.

In contrast, our training scheme does not require computationally expensive optimisation, and it always produces the correct correspondence between elements due to our key-value technique.

\begin{figure}
    \centering
    \includegraphics[width=0.85\linewidth]{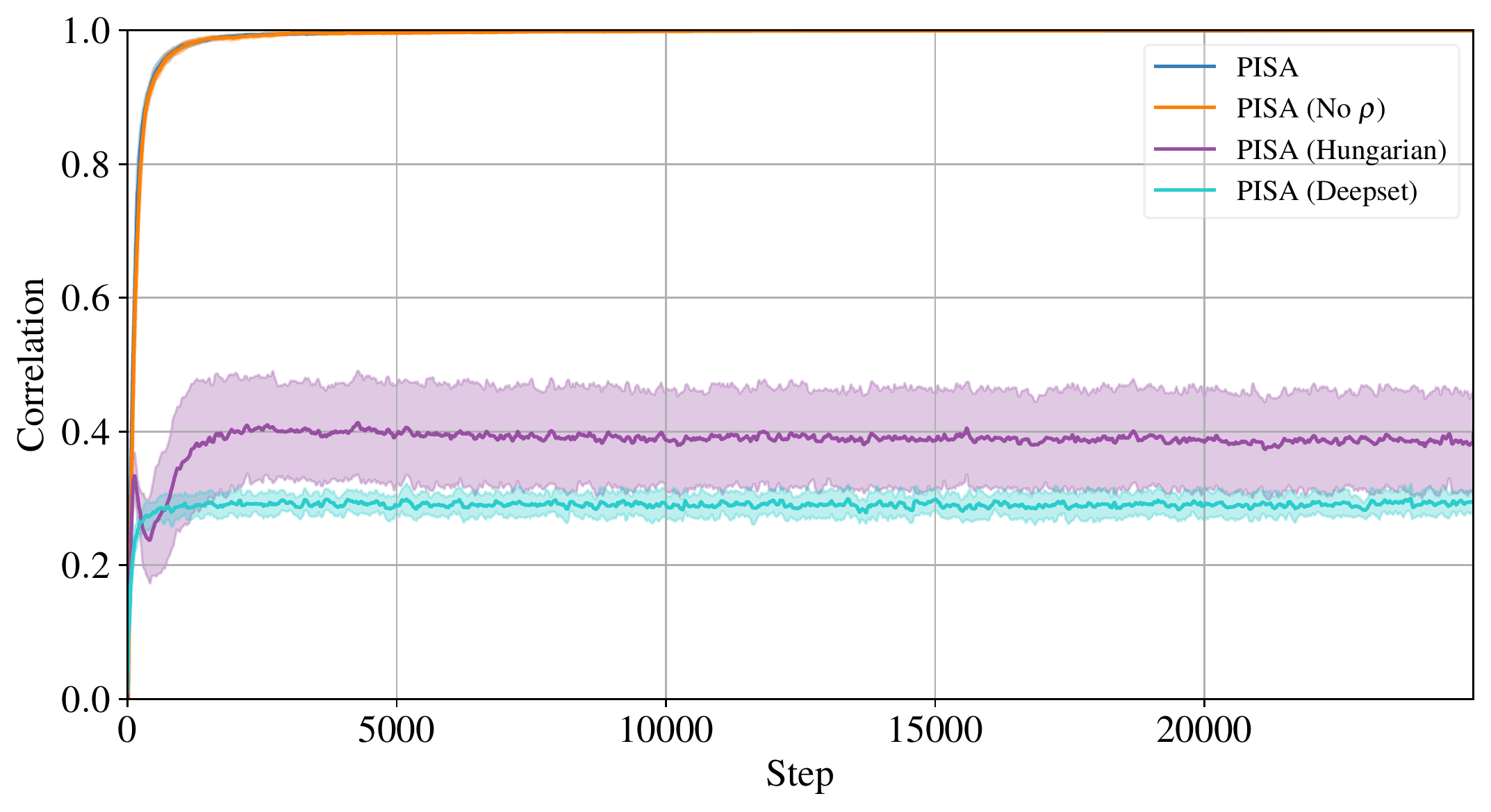}
    \caption{An ablation analysis over the components of PISA. \textit{PISA (No $\rho$)} denotes a version the model without $\rho$, whereby keys are assigned to values according to their input ordering. \textit{PISA (Hungarian)} refers to the same model as PISA, but trained with a Hungarian algorithm to match inputs with outputs, instead of using the ground truth correspondence. Lastly, \textit{PISA (Deepset)} represents a variant of PISA where the encoder is replaced by Deep Sets \cite{deepsets}. The metric used in this ablation analysis is the correlation coefficient between the inputs and outputs.}
    \label{fig:ablation}
\end{figure}

\begin{figure}
    \begin{subfigure}[c]{0.7\linewidth}
        \centering
        \includegraphics[width=\linewidth]{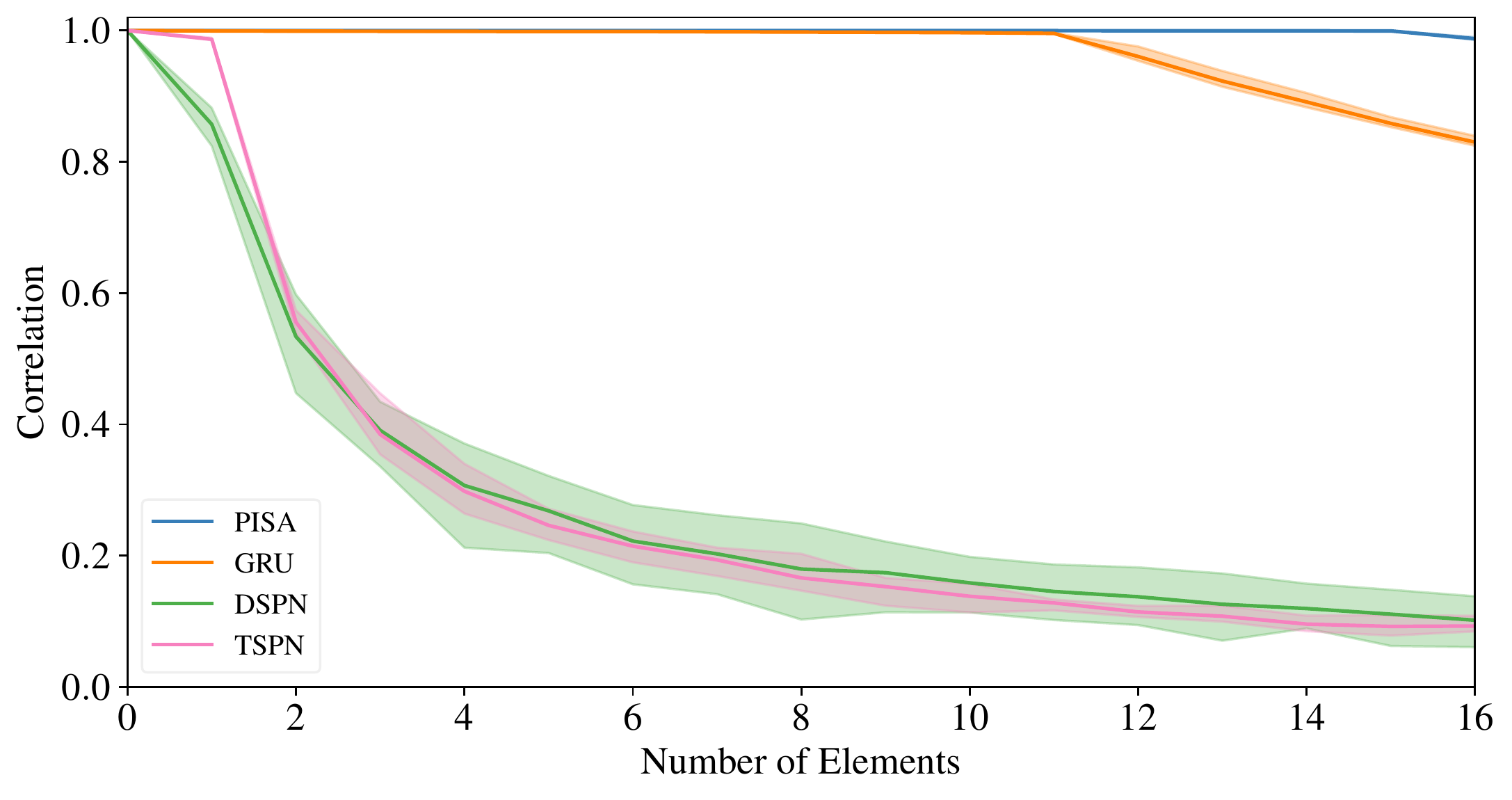}
        \caption{Correlation vs Number of Elements ($z \in \mathbb{R}^{96}$)}
    \end{subfigure}
    \begin{subfigure}[c]{0.7\linewidth}
        \centering
        \includegraphics[width=\linewidth]{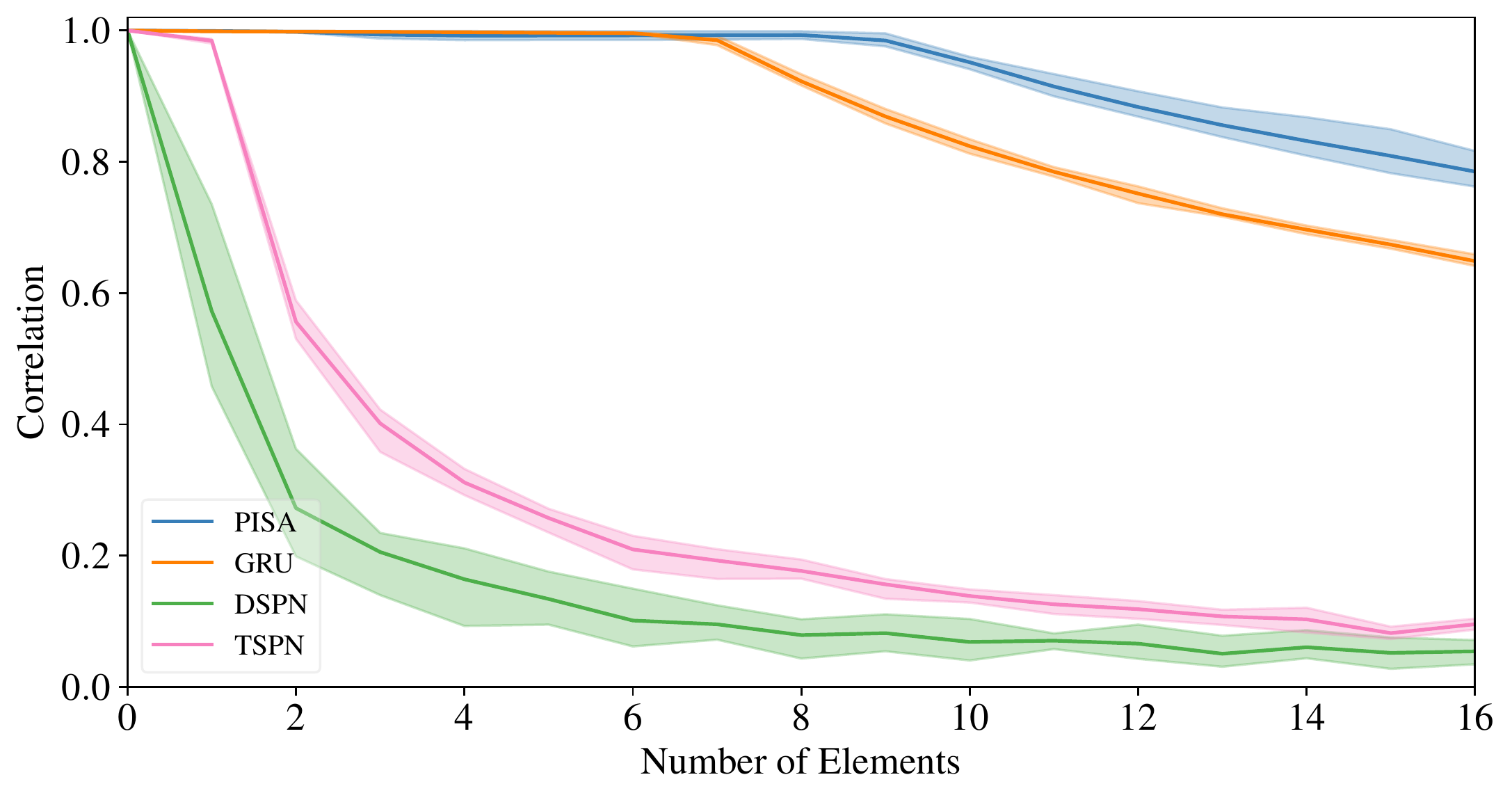}
        \caption{Correlation vs Number of Elements ($z \in \mathbb{R}^{48}$)}
    \end{subfigure}
    \caption{An analysis of the reconstruction accuracy (measured by correlation) for each method as a function of the number of elements in the set, evaluated with a latent of sufficient size ($z \in \mathbb{R}^{96}$) and under compression ($z \in \mathbb{R}^{48}$). Each line represents the mean over 1024 runs, and the shaded region denotes the min and max.}
    \label{fig:scale}
\end{figure}

\begin{figure}
    \centering
    \includegraphics[width=0.85\linewidth]{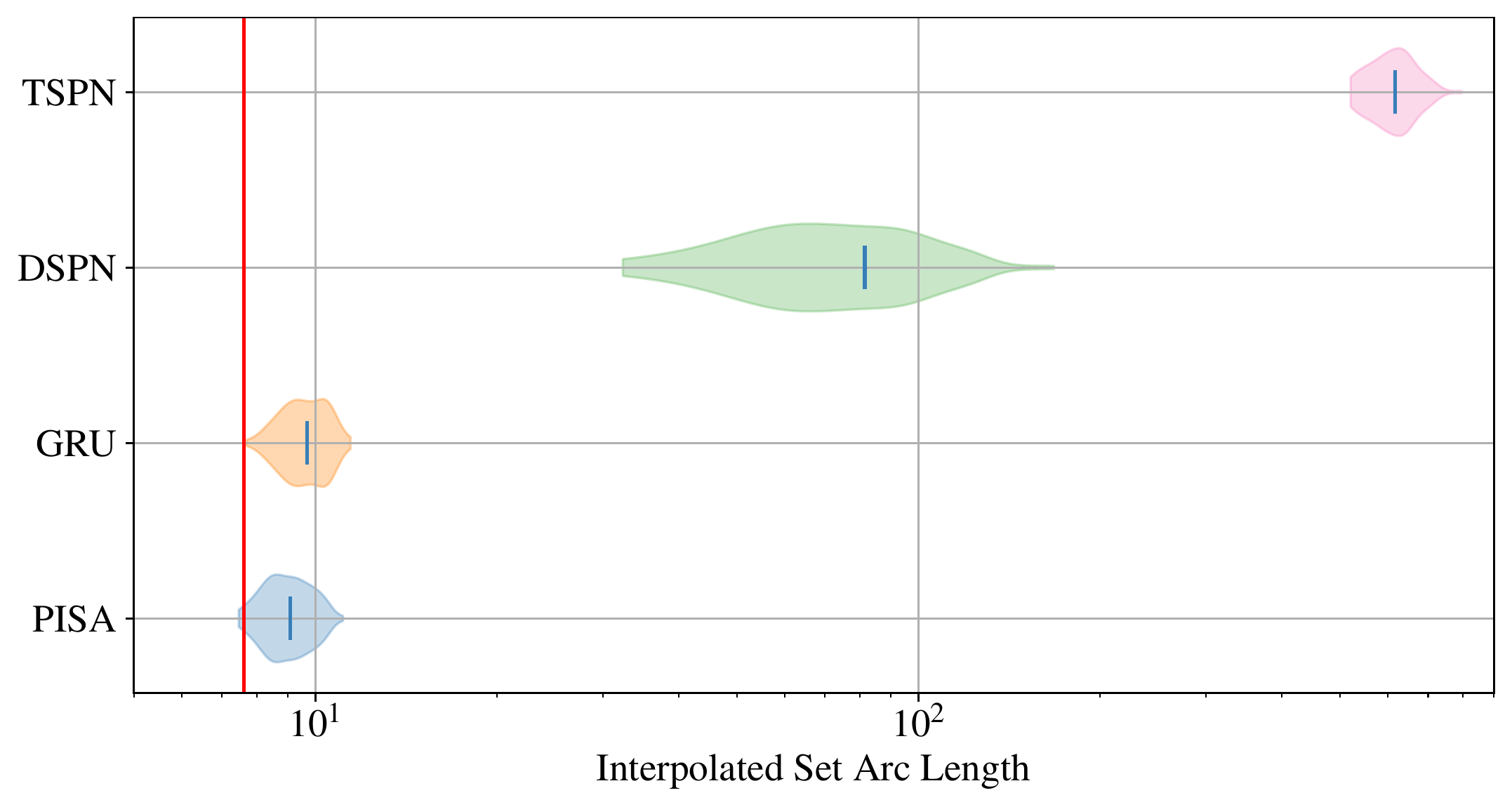}
    \caption{The integrated arc length travelled by all elements in the latent space interpolation experiment. This plot serves as a quantitative metric for the experiment shown in Fig. \ref{fig:interpolation}. Over 100 trials, we encode sets $\mathcal{X}_0$ and $\mathcal{X}_1$ into $z_0$ and $z_1$, interpolate between $z_0$ and $z_1$ in the embedding space, and decode the result: $\phi((1-\alpha) \cdot z_0 + \alpha \cdot z_1)$. We report the arc length of the reconstructed set produced by each method as it is interpolated: $\int^1_0 \left|\left| \frac{\partial \phi((1-\alpha) \cdot z_0 + \alpha \cdot z_1)}{\partial \alpha} \right|\right| \partial \alpha$. The red line represents the mean of the minimum possible arc lengths across all trials, calculated by finding the minimum cost assignment between the elements in sets $\mathcal{X}_0$ and $\mathcal{X}_1$ and computing the linear distance between each pair of elements.}
    \label{fig:dist}
\end{figure}

\begin{figure*}[t]
    \centering
    \includegraphics[width=0.85\textwidth]{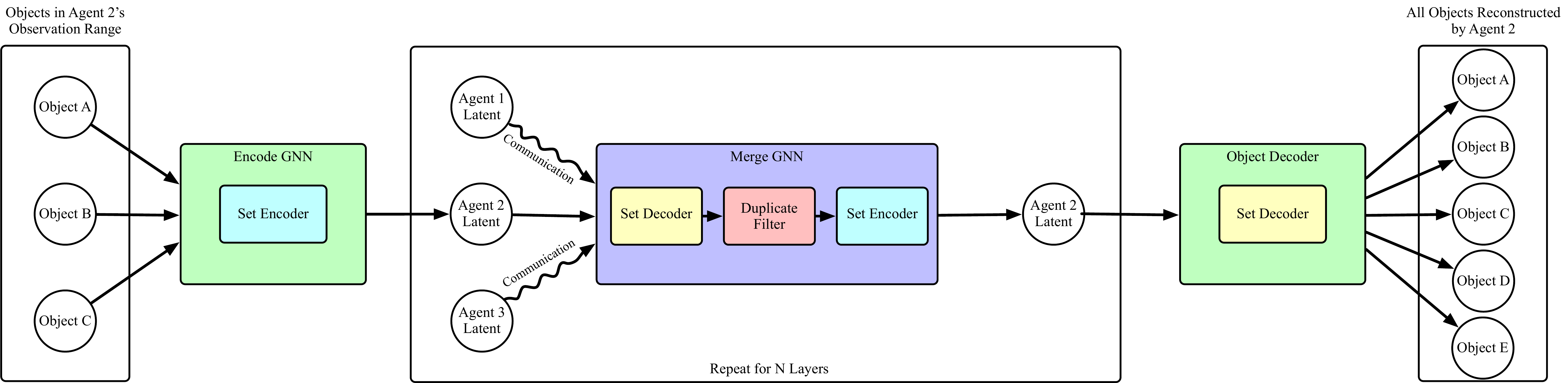}
    \caption{The Fusion GNN architecture. This schema is drawn from the perspective of agent 2, which can directly observe objects A, B, and C, and can communicate with agents 1 and 3. First, the local observation is encoded with the set encoder. Then, at each layer in the GNN, the agent decodes the latent states that it receives from its neighbours, filters out the duplicates, and encodes the resulting set. Finally, after the last GNN layer, the set decoder is applied to reconstruct the entire global observation.}
    \label{fig:fusion architecture}
\end{figure*}

\section{Experiments}

In this section, we first evaluate PISA against baselines, including GRU, DSPN, and TSPN. For a fair evaluation, we select a non-application-specific scenario: encoding and reconstructing a randomly generated set. Once we have demonstrated the performance of our method in comparison to baselines, we show its usefulness by applying it to a multi-agent problem: fusion of observations within a partially observable multi-agent system.

For all of the baselines, we use official code (Github repositories published by the original authors for DSPN and TSPN, and PyTorch's implementation of GRU), adapted to a common interface. All methods use their own recommended hyperparameters, with the exception of a common learning rate of $10^{-3}$ and latent state size (defined in the experiments). For DSPN and TSPN, we use Hungarian loss (as opposed to Chamfer loss). In DSPN, we run the decoder for $30$ iterations (the highest number of iterations used in the paper, which produced the highest fidelity outputs).

\subsection{Random Set Reconstruction}
\label{section: experiments random}

\textbf{Problem Formalisation.} In the random set reconstruction problem, the task is to compress a set $\mathcal{X}$ into a latent representation $z$, and then reconstruct it into $\hat{\mathcal{X}}$ such that the loss $\mathcal{L}_\mathrm{mse}(\mathcal{X}, \hat{\mathcal{X}})$ is minimised. Elements $x_i \in \mathbb{R}^6$ from each set are drawn from a normal distribution $\mathcal{N}(0_{6},I_{6})$, and the number of elements $n \in [0..16]$ is drawn from a discrete uniform distribution. Two latent sizes $d_z \in \{48, 96\}$ are tested to evaluate various degrees of compression. 

A solution to the random set reconstruction problem can be used as a permutation-invariant aggregation method which minimises the loss of data. This is extremely useful in the multi-agent domain, particularly in systems with GNN-based policies. For example, it can operate at the agent level as a method of combining messages from neighbouring agents, or at the global level as a graph pooling layer.

\textbf{Results.} In our first experiment, we compare PISA against baseline methods, evaluating both the mean squared error of the reconstruction and the correlation coefficient between the ground truth and reconstructed elements (Fig \ref{fig:sae results}). We run this experiment with hidden sizes of $96$ (which is theoretically just large enough to store all of the the inputs) and $48$ (which requires some compression). The results show that PISA outperforms all of the baselines in both scenarios, achieving a correlation coefficient of $0.9999$ for $z \in \mathbb{R}^{96}$ and $0.9180$ for $z \in \mathbb{R}^{48}$. The next best method, GRU, achieves a correlation of $0.9412$ for $z \in \mathbb{R}^{96}$ and $0.8228$ for $z \in \mathbb{R}^{48}$.

The conclusions we draw from the loss and correlation metrics are corroborated by the qualitative results in Fig \ref{fig:sae results}. We provide a visualisation of the reconstructed sets using each method, representing each $x_i \in \mathbb{R}^6$ datapoint as a circle with parameterised position, colour, and radius. The results show that PISA's reconstruction is near perfect, GRU's reconstruction has some error, and the reconstructions of DSPN and TSPN are very poor, making it difficult to distinguish the intended correspondences between inputs and outputs. We note that the performance of DSPN and TSPN for up to $2$ elements is satisfactory, but performance falls off sharply as the number of elements increases (Fig. \ref{fig:scale}).

\begin{figure}[h]
    \centering
    \includegraphics[width=0.85\linewidth]{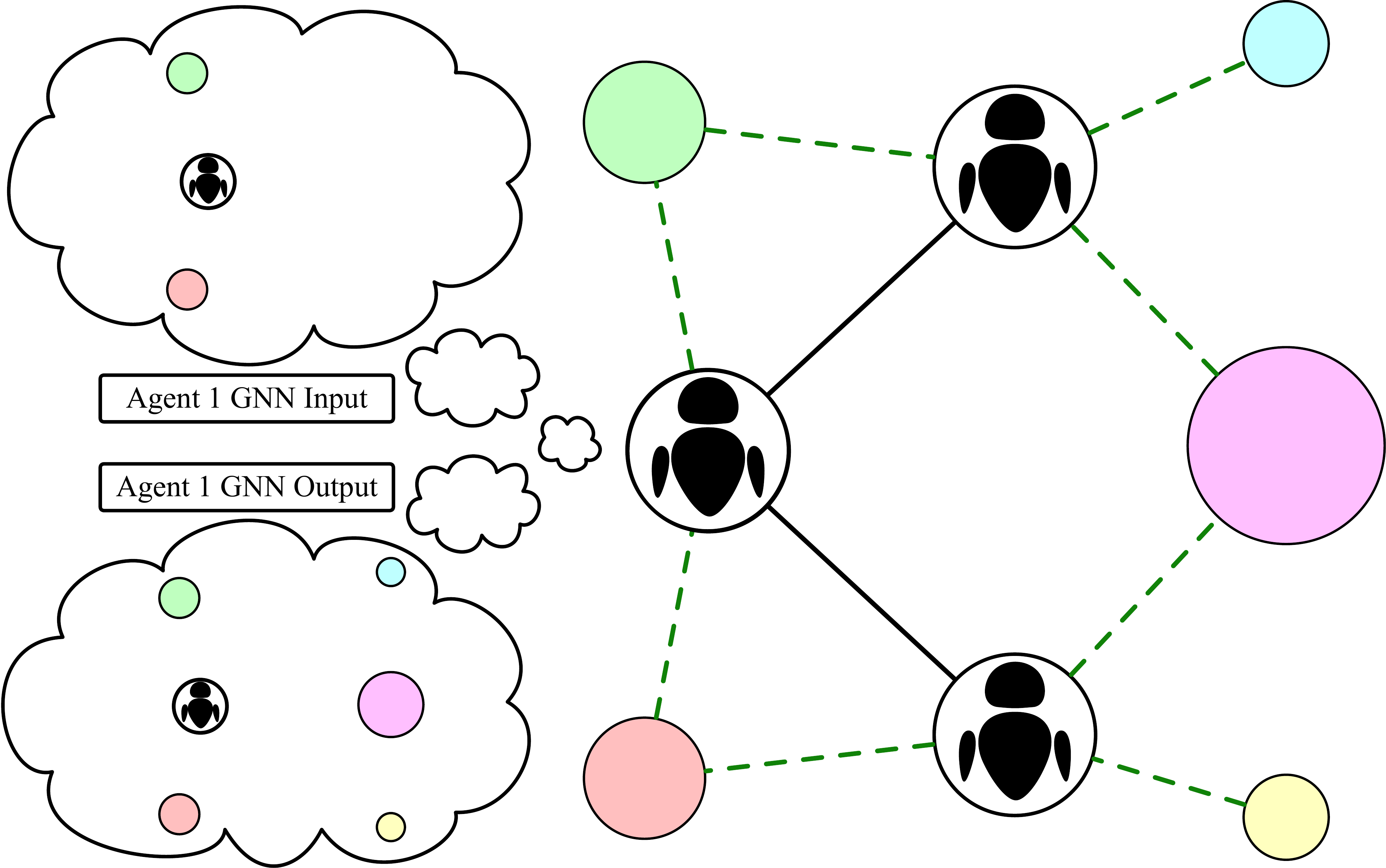}
    \caption{The Multi-Agent Sensor Fusion Problem. In this environment, there is a set of agents $\mathcal{A}$ (depicted as robots) and a set of objects $\mathcal{O}$ (depicted as coloured circles). The communication edges $\mathcal{E}^{\mathrm{(com)}}$ define the links between agents which can communicate (shown as black lines). The observation edges $\mathcal{E}^\mathrm{(obs)}$ define the objects which are in each agent's local observation range (shown as dotted green lines). The goal of the task is for each agent to use communication to reconstruct the entire global observation.}
    \label{fig:fusion problem}
\end{figure}

\textbf{Latent Space Sensitivity.} In addition to reconstruction loss, one extremely important property in autoencoders is the propensity to map similar elements to similar embeddings. The pursuit of this property forms the motivation for methods like variational autoencoders, which incentivise similarity preservation even at the cost of reconstruction error \cite{vae}. Similarity preservation is desirable because it leads to better stability and generalisation in networks using the embedding as an input.

Although neither our method nor the baselines specifically optimise for the sensitivity of the latent space, some methods lend themselves to more similarity-preserving embeddings than others (as a product of the architectures themselves). In this experiment, we analyse the effect of interpolating in the latent space (and, by extension, we test the property that similar inputs map to similar embeddings). We do this by encoding two random sets $\mathcal{X}_0$ and $\mathcal{X}_1$ into latents $z_0$ and $z_1$, interpolating between $z_0$ and $z_1$ in the latent space with a weighted average: $(1-\alpha) \cdot z_0 + \alpha \cdot z_1$ (for $\alpha$ from $0$ to $1$), and decoding the result.

As shown in Fig. \ref{fig:interpolation} and Fig. 
\ref{fig:dist}, PISA exhibits a similarity-preserving latent space. As we interpolate from $z_0$ to $z_1$, the decoded set smoothly transforms through logical intermediate sets. In contrast, DSPN and TSPN are both highly sensitive, so their decodings of the interpolated latent state are highly unstable. GRU is the only other method which demonstrates a smooth transition, and therefore low sensitivity to input noise. However, the intermediate states produced by GRU do not necessarily carry the meaning of a weighted average between $\mathcal{X}_0$ and $\mathcal{X}_1$---for example, note that while the visualisations of $\mathcal{X}_0$ and $\mathcal{X}_1$ both have large circles, the visualisation of $\hat{\mathcal{X}}_{0.5} = \frac{1}{2}z_0 + \frac{1}{2}z_1$ only has small circles.

\textbf{Ablation Analysis.} In addition to our comparison against baseline methods, we also run an ablation analysis to validate each component of our architecture (Fig. \ref{fig:ablation}). The results show that PISA performs significantly better than variants which (1) use the hungarian algorithm to match inputs and outputs in the loss function, and (2) use Deep Sets \cite{deepsets} in place of our encoder. On the other hand, the variant of PISA without the network $\rho$ to match keys with values (and therefore is not permutation-invariant) achieves the exact same performance as the base model. This is a significant result---our architecture is still able to generalise between different permutations without $\rho$. The purpose of the key-value matching criterion is simply to \textit{guarantee} permutation-invariance.

\begin{figure*}[t]
    \begin{subfigure}[b]{.31\textwidth}
    	\centering
            \includegraphics[width=0.99\textwidth]{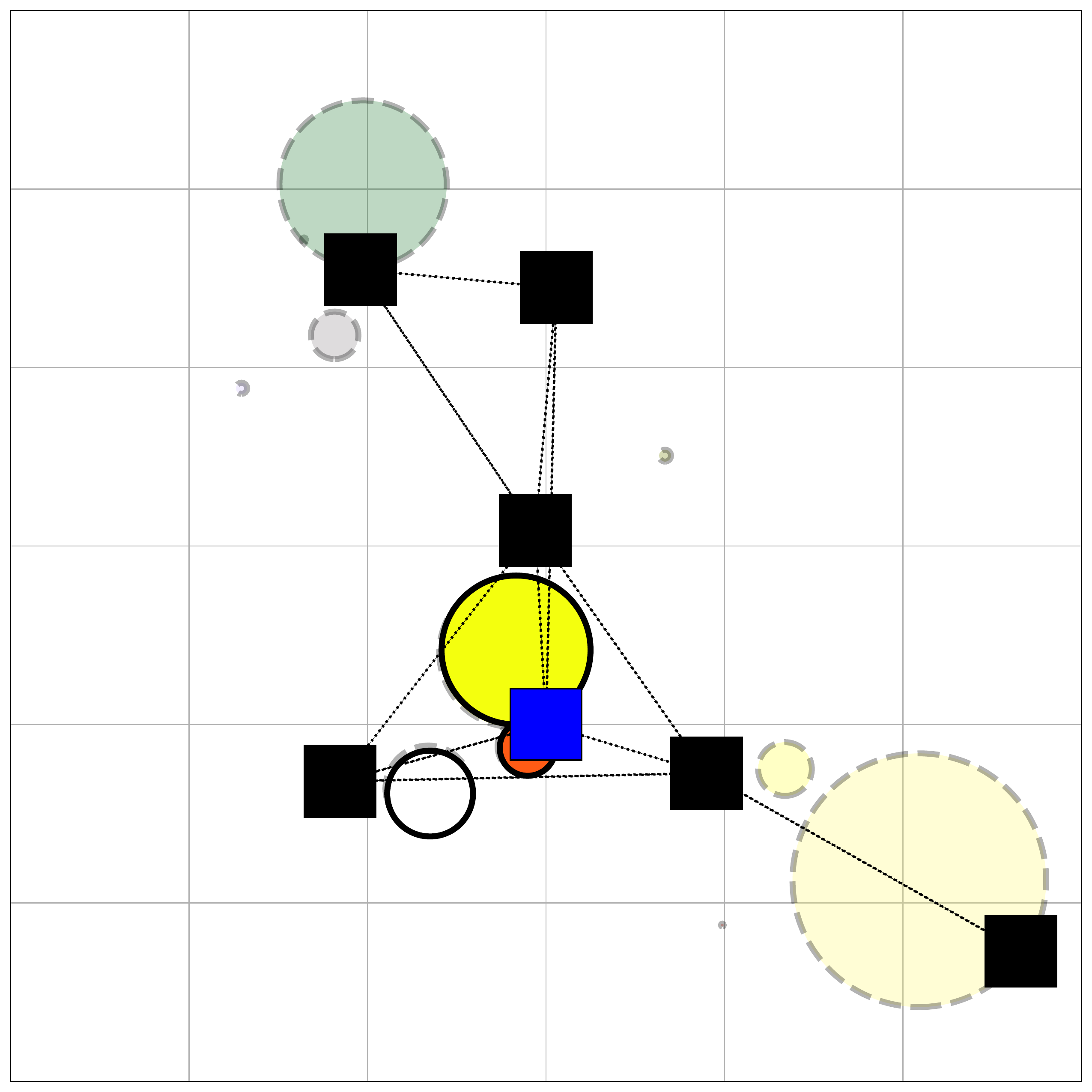}
    	\caption{Fusion GNN Layer 0}
    \end{subfigure}
    \hfill
    \begin{subfigure}[b]{.31\textwidth}
            \centering
            \includegraphics[width=0.99\textwidth]{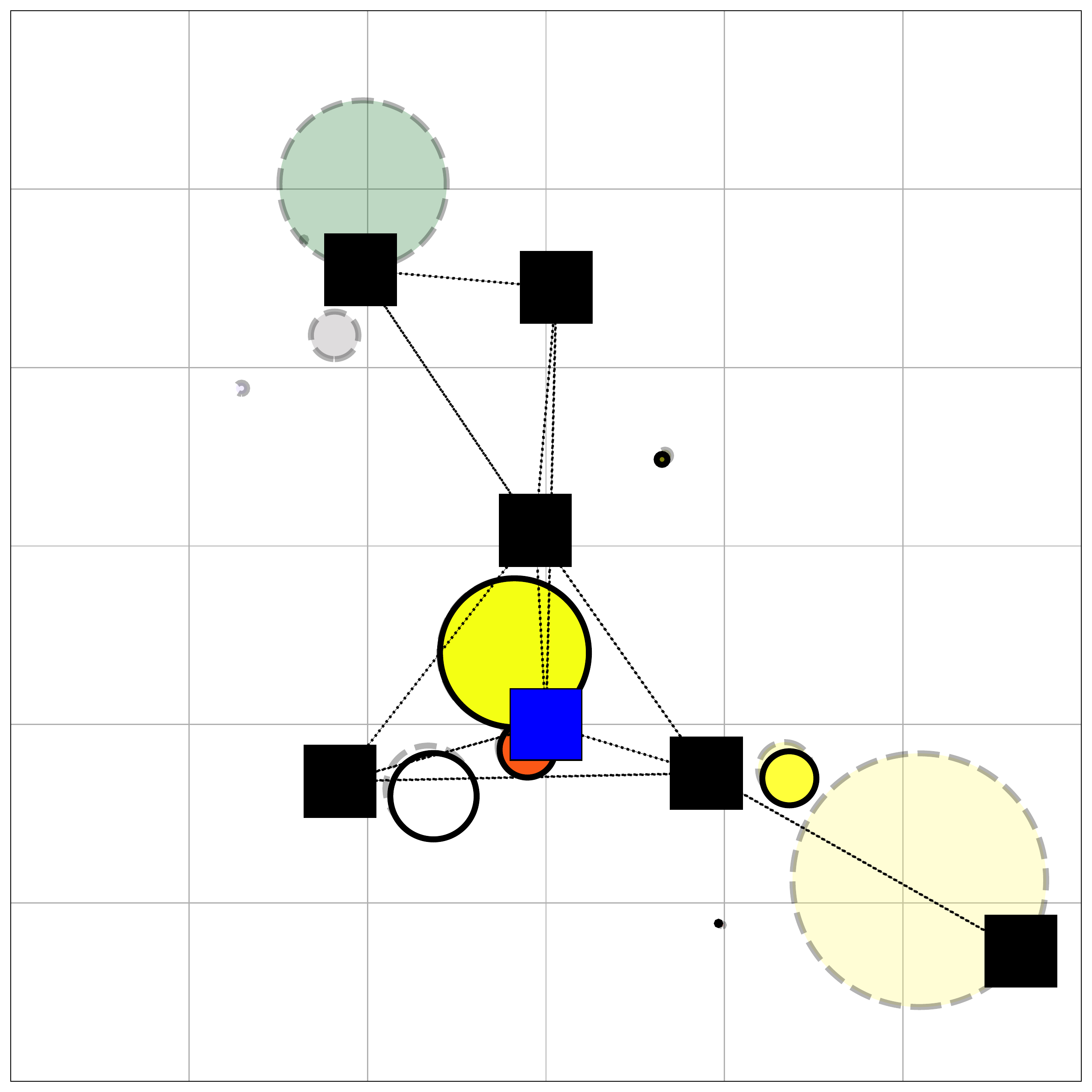}
    	\caption{Fusion GNN Layer 1}
    \end{subfigure}
    \hfill
    \begin{subfigure}[b]{.31\textwidth}
            \centering
            \includegraphics[width=0.99\textwidth]{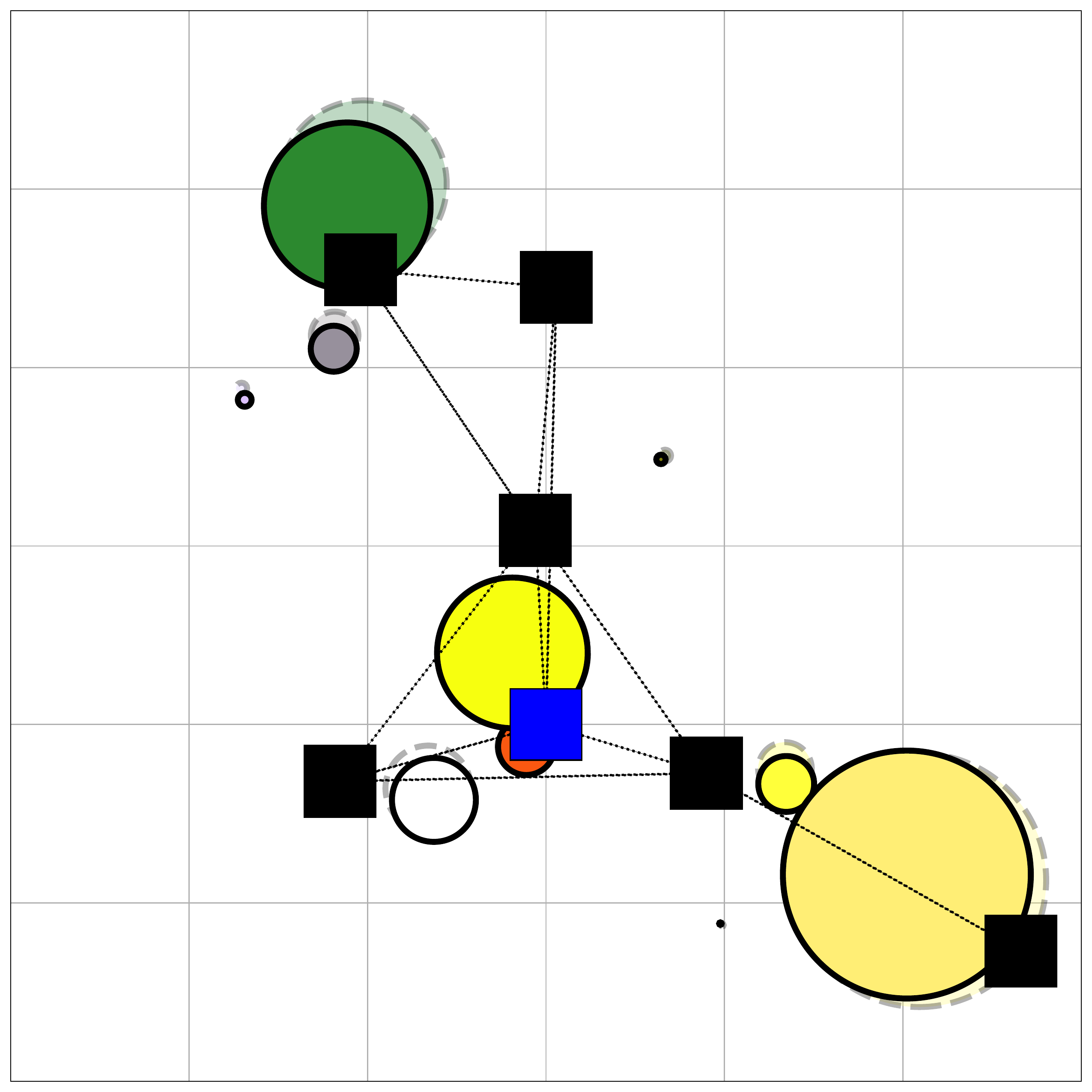}
    	\caption{Fusion GNN Layer 2}
    \end{subfigure}
    \caption{Evaluation of the trained Fusion model. In this visualisation, the $7$ agents are represented by black squares, and the $10$ observations are visualised as coloured circles. The plot is from the point of view of agent $0$, depicted as a blue square. The transparent circles represent the ground truth global observation, while the circles with full opacity represent the scene reconstructed by agent $0$. At layer $0$ of the GNN, agent $0$ only has access to its own observations. However, after multiple rounds of message passing, agent $0$ can reconstruct objects outside of its observation range.}
    \label{fig:fusion results}
\end{figure*}

\subsection{Multi-Agent Sensor Fusion}

\textbf{Problem Formalisation.} The multi-agent sensor fusion problem concerns the task of using communication to combine sets of observations within a partially observable multi-agent system, producing global observability (assuming the graph of communication links between agents is connected). Consider a graph $\mathcal{G}_\mathrm{com} = \langle \mathcal{A}, \mathcal{E}^{\mathrm{(com)}} \rangle$ consisting of a set of agents $\mathcal{A} \equiv [1..n]$ and and edges $\mathcal{E}^\mathrm{(com)}$, where the edge $e_{i,j}^\mathrm{(com)}$ represents a communication link between agents $i$ and $j$, and $\mathcal{E}_i^{\mathrm{(com)}} = \{ j \mid e_{i,j}^\mathrm{(com)} \in \mathcal{E}^\mathrm{(com)} \}$ represents the set of agents in communication range of agent $i$. Let $\mathcal{G}_\mathrm{obs} = \langle \mathcal{A} \cup \mathcal{O}, \mathcal{E}^\mathrm{(obs)} \rangle$ be a bipartite graph representing the objects which can be observed by each agent, where $\mathcal{O}$ is the set of objects, and $\mathcal{E}^\mathrm{(obs)}$ is the set of edges between agents and objects. Each object $o_i \in \mathcal{O}$ contains information $o_i \in \mathbb{R}^6$ which is provided to the agents as an observation. Without communication, each agent can observe the set of objects $\mathcal{E}_i^{\mathrm{(obs)}} = \{ o_j \mid e_{i,j}^{\mathrm{(obs)}} \in \mathcal{E}^\mathrm{(obs)} \}$. However, with communication, agents can gain access to information about objects which they cannot directly observe. If $\mathcal{X}_i^{(l)}$ is the set of objects that agent $i$ has reconstructed after $l$ rounds of communication, then the goal of this task is to learn a communication strategy with fixed-size messages such that $\lim_{l \rightarrow \infty} \mathcal{X}_i^{(l)} = \mathcal{O}$.

Since this communication strategy is trained with an autoencoder, it is application-agnostic. That is, the communication strategy can be trained on synthetic data, and then that pre-trained submodule can be deployed in multiple disparate systems. It also allows the number of layers to be updated dynamically at execution time, enabling systems to make the number of communication steps arbitrarily large or small in order to receive all necessary information (\textit{i.e} the maximum of the lengths of the shortest paths between every pair of agents in $\mathcal{G}^\mathrm{(com)}$).

One potential application for this is multi-agent reinforcement learning. Even in single-agent reinforcement learning, the task of learning an appropriate representation for the state is a major bottleneck with respect to sample efficiency \cite{learnlatent}. That is, it is difficult to learn an encoder with only a reward signal. This problem is only exacerbated when moving to the multi-agent domain, wherein the size of the state space is even larger. However, with a multi-agent sensor fusion system, the task of learning an encoder and communication scheme can be pre-trained with unsupervised learning, allowing the reinforcement learning stage of training to focus on learning the task. In problems with local rewards, this allows the multi-agent reinforcement learning problem to be reformulated into a single-agent reinforcement learning problem, where the latent state from the communication system is the observation. This allows the problem to be solved by single-agent reinforcement learning methods.

Another possible application for the fusion model is hierarchical reinforcement learning in a multi-agent system \cite{HMARL}. Since communication in MARL is usually trained alongside a policy to complete a given task \cite{qingbiaognn}, agents operating under different policies cannot communicate with each other. This is a problem in hierarchical reinforcement learning, where the team must transition between policies, and different subsets of agents might select different policies at any given timestep. However, with our fusion model, a generalised communication scheme can be trained for all of the policies.

\noindent \textbf{Model.} Our multi-agent sensor fusion model is defined by a novel GNN architecture. The architecture has three subcomponents: a set encoder $\psi$, a set decoder $\phi$, and a duplicate filter $f$. The set encoder compresses a set of objects into a message which can be passed to other agents:
\begin{equation}
z_i^{(l)} = \psi(\mathcal{X}_i^{(l)}).
\end{equation}
Similarly, the set decoder reconstructs a set of objects given a message: 
\begin{equation}
\mathcal{X}_i^{(l+1)} = \phi(z_i^{(l)}).
\end{equation}
Finally, the filter takes in a \textit{set} of sets of objects, and returns the union over those sets (by identifying and removing the duplicates):
\begin{equation}
\mathcal{X}_1 \cup \mathcal{X}_2 \cup \hdots \mathcal{X}_n = f(\{\mathcal{X}_1, \mathcal{X}_2, \hdots, \mathcal{X}_n\}) .
\end{equation}
The filter $f$ is defined by training a binary classifier $g$ which determines if two reconstructed objects are the same $g: \mathcal{O} \times \mathcal{O} \mapsto [0,1]$, and evaluating it on every pair of objects.

The entire GNN is constructed with these three components, with parameter sharing across layers. At each layer, the GNN decodes the latent states from all of the neighbours, filters the duplicates, and encodes the result:
\begin{equation}
z_i^{(l+1)} = \phi \left( f \left( \left\{ \psi(z_j^{(l)}) \; \Bigr| \; j \in \mathcal{E}_i^{\mathrm{(com)}} \right\} \right) \right) .
\end{equation}
The loss for the GNN is calculated per-layer. The encoder and decoder are trained with the standard set autoencoder loss, which is applied between the input to the set encoder and the reconstructed output of the set decoder in the next layer. The filter is trained with a crossentropy loss, using a dataset of pairs of reconstructed observations at each layer $(\hat{o}_i, \hat{o}_j)$, and a label which indicates if those were reconstructed from the same original object $1(i=j)$.

\noindent \textbf{Results.} As shown in Fig. \ref{fig:fusion results}, our fusion model can successfully learn a communication scheme which allows agents to gain full observability of a system through message passing. The trained model's set autoencoder achieves a reconstruction correlation coefficient of $0.9999$, and the duplicate filter achieves an accuracy of $1.0$. 

While these quantitative results produce reconstructions which should be sufficient for most mutli-agent applications, we note that the elements slightly drift as the number of layers of the GNN increases, as a result of compounded error. Fortunately, there is a way that this effect could be mitigated in future work. Currently, all elements are re-encoded at every layer in the GNN, and if there are duplicates then there is no mechanism for choosing one element over another. However, since our set autoencoder architecture enables the addition and removal of elements after a set is initially encoded, it is possible to design the system in a manner such that only new elements are added. This would not only remove the error due to re-encoding, but also place a bias on choosing elements which come from fewer communication steps away when duplicates exist.

\section{Discussion}

In our experiments, we demonstrate that PISA is capable of producing near-perfect reconstructions and possesses a similarity-preserving latent space. The combination of these properties enable PISA to be used as a pre-trained submodule within a larger learned model. The encoder can be used as a permutation-invariant aggregator which minimises the loss of information, while the decoder can be used as a module for predicting variable-sized outputs given a vector of logits. Furthermore, the Fusion GNN can be used as a task-agnostic communication scheme for observations of a given size. Given the application-agnostic nature of our experiments, the results should hold when applied to any domain.

The primary attribute that sets our method apart from the baselines is our approach to the core problem in set autoencoding: producing a variable number of unique outputs from a single latent state. While our method solves this issue in a simple manner with a key-value approach, the baselines introduce problems with over-engineered workarounds. In the case of GRU, the primary issue is that it encodes and decodes sequentially, thereby including the ordering of the inputs in the embedding, and removing its ability to generalise across all orderings. This not only injects some instability (whereby the manner in which elements are processed depends on previous elements in the sequence), but also introduces non-permutation-invariance. For DSPN and TSPN, the primary issue is likely the fact that there is no correspondence between inputs and outputs. In order to apply a loss function, DSPN and TSPN use the Hungarian algorithm to match predicted elements with the closest elements in the input set. However, this matching process is imperfect, and consequently introduces some noise into the loss signal which makes it more difficult to learn (as shown in our ablation analysis in Fig. \ref{fig:ablation}). In the DSPN and TSPN papers, the primary experiment for evaluation was autoencoding point clouds forming MNIST digits. In those experiments, if the matching is imperfect, it can still produce the desirable macroscopic behaviour (even if individual elements are in the wrong places).

One unique property of our model which we do not explore in this paper is its ability to perform transforms on the latent space. Unlike in other methods, sets can be combined simply by adding embeddings---there is no need to re-encode a new set. To insert an element $e$ to an existing embedding $z_0$ of size $n$, it must be encoded with key $n+1$ to produce embedding $z_e$. Then, the two embeddings can be combined by adding them: $z_0 + z_e$. Removal of elements through subtraction is also possible, but it requires the user to keep track of keys (to identify the element which will be removed, and to remember which keys are still present after a subset is removed). 

Although we focus on multi-agent systems in this paper, PISA has applications in many different domains. The encoder can be used in any application which involves reasoning over graphs, including natural language processing \cite{nlp}, image processing \cite{nlnn}, and knowledge graphs \cite{knowledge}. The decoder can be used in neural cellular automata on graphs (expanding sets of children nodes), prediction of scenes with multiple objects (either predicting point clouds or semantic objects), and multimodal continuous action distributions (where a vector of logits is mapped to a set of Gaussians which are composed together). Given the experimental success of PISA, we hope that it will open up future avenues of research in these fields.



\section{Conclusion}

In our experiments, we have demonstrated that PISA outperforms all baselines on a reconstruction task with random data. Our model produces near-perfect reconstructions up to a compression factor of $1.0$, and significantly better reconstructions than the baselines with more compression. Furthermore, it possesses a well-behaved latent space, mapping similar inputs to similar embeddings. In addition to benchmarking the performance of PISA on the reconstruction of random data, we demonstrate its usefulness in the multi-agent sensor fusion problem---our novel GNN architecture uses PISA to define a generalised communication strategy. However, this represents just one of many possible applications. Given its success with completely random data, it is possible for PISA to be used as a submodule in any application which involves learning over set or graph-structured data.

\section*{Acknowledgements}
\small
Ryan Kortvelesy is supported by Nokia Bell Labs through their donation for the Centre of Mobile, Wearable Systems and Augmented Intelligence to the University of Cambridge. A. Prorok acknowledges funding through ERC Project 949940 (gAIa).

\balance
\bibliographystyle{ACM-Reference-Format}
\bibliography{ref}

\newpage
\nobalance

\appendix
\section*{Appendix}
\section{Additivity}
\label{appendix: additivity}

This stems from the fact that the last operation in our encoder is a summation, and therefore can be split via the associative property:

\begin{align}
\begin{split}
    \psi(\mathcal{X}_0 \cup \mathcal{X}_1) = & \sum_{x \in \mathcal{X}_0}^n \left[ \psi_\mathrm{key}(\rho(x)) \odot \psi_\mathrm{val}(x) \right] \\
    + & \sum_{x \in \mathcal{X}_1}^n \left[ \psi_\mathrm{key}(\rho(x)) \odot \psi_\mathrm{val}(x) \right] \\
    + & \;\; \lambda_\mathrm{enc}(|\mathcal{X}_0 \cup \mathcal{X}_1|)
\end{split}
\end{align}

Furthermore, since $\lambda_\mathrm{enc}$ is linear, we can use the property of linearity to split it (for disjoint sets $\mathcal{X}_0$ and $\mathcal{X}_1$): 

\begin{equation}
    \lambda_\mathrm{enc}(|\mathcal{X}_0 \cup \mathcal{X}_1|) = \lambda_\mathrm{enc}(|\mathcal{X}_0|) + \lambda_\mathrm{enc}(|\mathcal{X}_1|)
\end{equation}

The resulting equation can be grouped into expressions which depend only on $\mathcal{X}_0$ or $\mathcal{X}_1$: 

\begin{align}
\begin{split}
    \psi(\mathcal{X}_0 \cup \mathcal{X}_1) = & \sum_{x \in \mathcal{X}_0}^n \left[ \psi_\mathrm{key}(\rho(x)) \odot \psi_\mathrm{val}(x) \right]  + \lambda_\mathrm{enc}(|\mathcal{X}_0|) \\
    + & \sum_{x \in \mathcal{X}_1}^n \left[ \psi_\mathrm{key}(\rho(x)) \odot \psi_\mathrm{val}(x) \right] + \lambda_\mathrm{enc}(|\mathcal{X}_1|) 
\end{split}
\end{align}

Note that these expressions can be replaced with $\psi(\mathcal{X}_0)$ and $\psi(\mathcal{X}_0)$, so we can show that our model possesses the property of additivity: $\psi(\mathcal{X}_0 \cup \mathcal{X}_1) = \psi(\mathcal{X}_0) + \psi(\mathcal{X}_1)$. One caveat to this property is that one of the sets must be encoded using keys with an offset (using keys from $|\mathcal{X}_0|+1$ to $|\mathcal{X}_0|+|\mathcal{X}_1|$ instead of $1$ to $|\mathcal{X}_1|$). Otherwise, multiple values would be associated with each key, so decoding would not work. However, if the sets are encoded normally, then addition in the latent space \textit{can} be used for interpolation---we use this property for our experiments in section \ref{section: experiments random}.

\begin{figure*}[b]
    \centering
    \begin{subfigure}[c]{0.99\linewidth}
        \centering
        \includegraphics[width=0.99\textwidth]{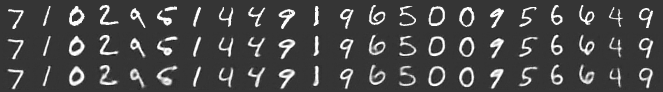}
        \caption{$\frac{1}{2}$ Set Autoencoder Compression Ratio}
        \label{fig:0.5comp}
    \end{subfigure}
    \begin{subfigure}[c]{0.99\linewidth}
        \centering
        \includegraphics[width=0.99\textwidth]{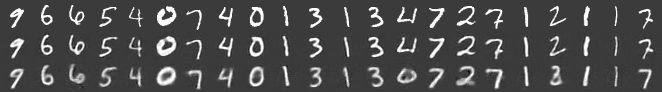}
        \caption{$\frac{1}{4}$ Set Autoencoder Compression Ratio}
        \label{fig:0.25comp}
    \end{subfigure}
    \caption{Results for MNIST set reconstruction, where sets of up to $16$ images are encoded and reconstructed. The outer level of our model is a CNN autoencoder which compresses $32 \times 32$ images into a latent of size $64$. The inner level is our set autoencoder, which compresses to either $\frac{1}{2}$ the maximum set size, as shown in (\subref{fig:0.5comp}) or $\frac{1}{4}$ the maximum set size, as shown in (\subref{fig:0.25comp}). The numbers in the top row are the ground truth, those in the middle row are reconstructed by the CNN, and those in the bottom row are reconstructed by the entire CNN/PISA architecture (where set sizes range between $1$ and $16$).}
    \label{fig:mnist}
\end{figure*}

\section{MNIST}
\label{appendix: mnist}

In this section we evaluate our set autoencoder on an MNIST dataset. While the rest of our paper uses randomly generated data, this experiment deals with structured data. Furthermore, it shows that our set autoencoder works as a subcomponent in a larger neural network architecture. The full model in this experiment consists of a CNN encoder, set encoder, set decoder, and then a CNN decoder. The CNN encoder compresses $\mathbb{R}^{32 \times 32}$ images into a latent with size $\mathbb{R}^{64}$, and the set encoder compresses up to $16$ of these elements by a compression factor of $\frac{1}{2}$ the max size $\mathbb{R}^{\frac{1}{2} \cdot 64 \cdot 16}$ or $\frac{1}{4}$ the max size $\mathbb{R}^{\frac{1}{4} \cdot 64 \cdot 16}$.

The results are shown in Fig. \ref{fig:mnist}. With a compression factor of $\frac{1}{2}$, the model generates near-perfect reconstructions. However, with a compression factor of $\frac{1}{4}$, the model makes some errors. As shown in Fig. \ref{fig:0.25comp}, the model incorrectly predicts a $4$ as a $0$ and a $2$ as a $3$. Interestingly, some of the other numbers are correct, but not exact matches to the ground truth. For example, one of the digits in the ground truth is a $0$ which is not closed, but the predicted digit closes the loop. Another ground truth digit is a $7$ with a cross through the middle, but the prediction omits the cross. These errors indicate that under compression, the set autoencoder attempts to preserve the most semantically important information. That is, the model has learned from examples that $0$ \textit{should} be a closed loop, and $7$ often does not have a cross through it. So, while under compression, the model maximises its reconstruction accuracy by storing enough information to get close (primarily by encoding \textit{which} digit it is), and discarding information which can often be inferred.

\end{document}